\documentclass[10pt,journal,compsoc]{IEEEtran}
\usepackage[nocompress]{cite}
\usepackage{graphicx}
\usepackage{caption} 
\usepackage{subcaption}
\usepackage{amsmath}
\usepackage{mathtools}
\usepackage{amsfonts}
\usepackage{amssymb}
\usepackage{todonotes}
\usepackage{cancel}
\usepackage{algorithm}
\usepackage{algpseudocode}
\usepackage{multirow} %
\usepackage{cuted}
\usepackage{multicol}
\usepackage{adjustbox}
\usepackage{booktabs} 
\usepackage{bm}
\usepackage{hyperref} 
\definecolor{mydarkblue}{rgb}{0,0.08,0.45}
\definecolor{myfavblue}{rgb}{0.1176, 0.392, 1.0}
\hypersetup{
    colorlinks=true,
    linkcolor=mydarkblue,
    citecolor=mydarkblue,
    filecolor=mydarkblue,
    urlcolor=mydarkblue}

\newcommand\numberthis{\addtocounter{equation}{1}\tag{\theequation}}
\begin{document}

\title{Sampling-Free Probabilistic Deep State-Space Models}

\author{Andreas~Look,
        Melih~Kandemir,
        Barbara~Rakitsch,
        and~Jan~Peters,~\IEEEmembership{Fellow,~IEEE}
\IEEEcompsocitemizethanks{\IEEEcompsocthanksitem A. Look and B. Rakitsch  are with the Bosch Center for Artificial Intelligence, Renningen, Germany.\\
E-mail: \{andreas.look, barbara.rakitsch\}@bosch.com
\IEEEcompsocthanksitem M. Kandemir is with University of Southern Denmark, Odense, Denmark. \\
E-mail: kandemir@imada.sdu.dk. 
\IEEEcompsocthanksitem J. Peters is with Intelligent Autonomous Systems Group, Technical University Darmstadt, Darmstadt, Germany, and also with the Max Planck Institute for Intelligent Systems, 72076 Tübingen, Germany. \\
E-mail: peters@ias.informatik.tu-darmstadt.de.}
}

\IEEEtitleabstractindextext{%

\begin{abstract}
Many real-world dynamical systems can be described as \textit{State-Space Models} (SSMs). In this formulation, each observation is emitted by a latent state, which follows first-order Markovian dynamics. 
A \textit{Probabilistic Deep SSM} (ProDSSM) generalizes this framework to dynamical systems of unknown parametric form, where the transition and emission models are described by neural networks with uncertain weights. 
In this work, we propose the first deterministic inference algorithm for models of this type. Our framework allows efficient approximations for training and testing. 
We demonstrate in our experiments that our new method can be employed for a variety of tasks and enjoys a superior balance between predictive performance and computational budget. 
\end{abstract}}

\maketitle
\IEEEdisplaynontitleabstractindextext
\IEEEpeerreviewmaketitle

\IEEEraisesectionheading{\section{Introduction}\label{sec:introduction}}

\IEEEPARstart{M}{odeling} unknown dynamics from data is challenging, as it requires accounting for both the intrinsic uncertainty of the underlying process and the uncertainty over the model parameters. Parameter uncertainty, or epistemic uncertainty, is necessary to address the uncertainty arising from incomplete data. Intrinsic uncertainty, also known as aleatoric uncertainty, is essential to represent the inherent stochasticity of the system \cite{kendall2017uncertainties, depeweg2017_decomposition}.

Deep state-space models \cite{archer2015black, karl2016deep, krishnan17_dks} offer a principled solution for modeling the intrinsic uncertainty of an unidentified dynamical process.
At their core, they assign a latent variable to each data point, which represents the underlying state and changes over time while considering uncertainties in both observations and state transitions. 
Neural networks with deterministic weights describe the nonlinear relationships between latent states and observations.
Despite offering considerable model flexibility, these deterministic weights ultimately limit the models' ability to capture epistemic uncertainty.

On the other hand, most prior works that take weight uncertainty into account make either the simplifying assumption that the transition dynamics are noiseless \cite{yildiz2019_ode2vae, dandekar2020bayesian, iakovlev2022latent} or that the dynamics are fully observed \cite{depeweg2017_decomposition, depeweg2016learning}.
Both assumptions are not satisfied by many real-world applications and can lead to miscalibrated uncertainties.

There also exists a large body of work for state-space models \cite{ialongo19_vcdt, doerr18_prism, lindinger22_laplacegp}  that use Gaussian Processes to model state transition kernels instead of probabilistic neural networks.
While these methods respect both sources of uncertainty,
they do not scale well with the size of the latent space.
Finally, there is the notable exception of \cite{haussmann2021_pacsde} that aims at learning deep dynamical systems that respect both sources of uncertainty jointly. 
However, this approach requires to marginalize over the latent temporal states and the neural network weights via plain Monte Carlo, which is infeasible for noisy transition dynamics.

\begin{figure}[t]
	 \begin{subfigure}[t]{.5\columnwidth}
   \centering
\includegraphics[height=.8\columnwidth]{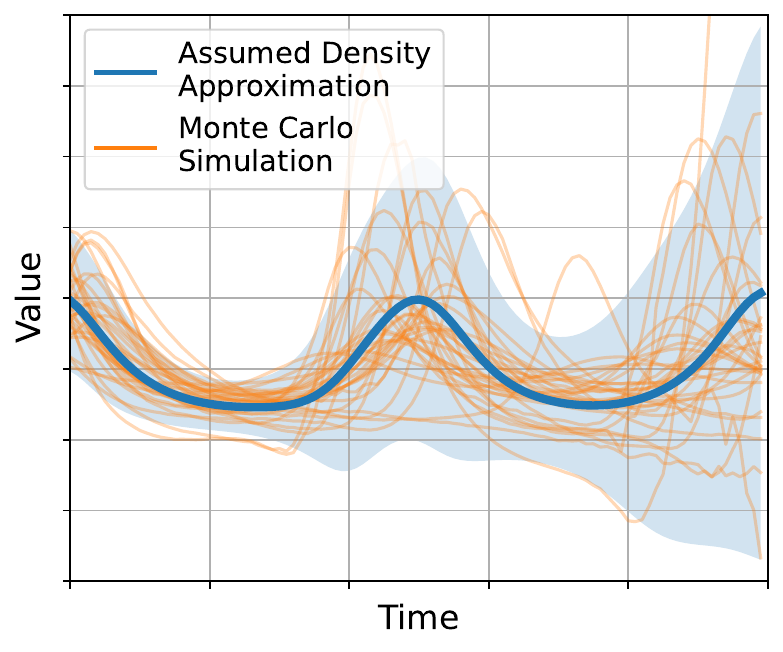}
	\caption{Uncertainty propagation.}\label{fig:main_1a}
	\end{subfigure}%
 \begin{subfigure}[t]{.5\columnwidth}
   \centering
\includegraphics[height=.8\columnwidth]{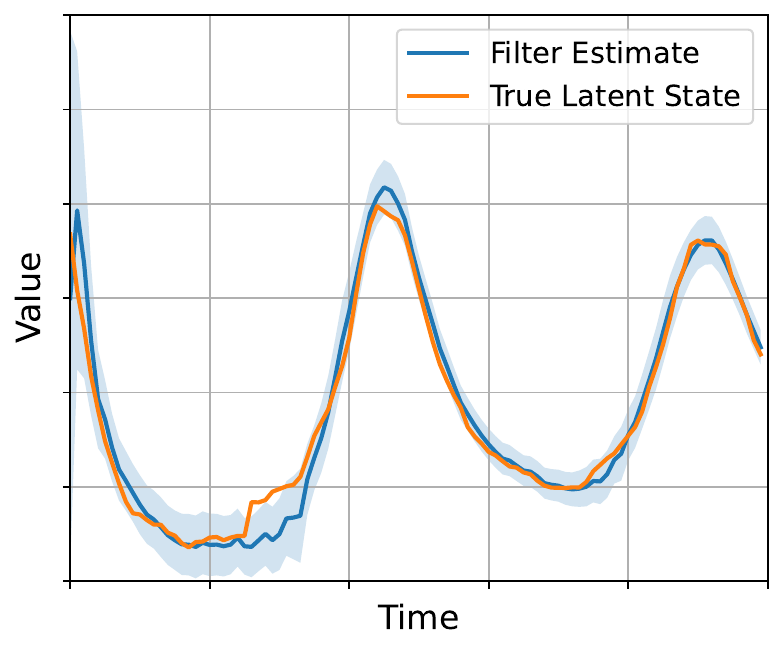}
\caption{Filtering.}\label{fig:main_1b}
	\end{subfigure}%
 \caption{
 We simulate a dynamical system  $p(x_{t+1}|x_t, w_t)$ with uncertainty over the weights $w_t \sim p(w_t)$. 
Our deterministic approximation scheme is shown in blue, where the solid line depicts the mean and the shaded area is the 95\% confidence interval.
 In Panel (a), we compare our approach with Monte Carlo (orange) for multi-step ahead predictions.
 Our deterministic approximation accurately captures the first two moments of the Monte Carlo generated samples.
In Panel (b), we move the dynamical system to a latent space and introduce an emission function $p(y_t|x_t)$. We compare our filtering distribution with the true latent state (orange).
The true latent trajectory lies within the 95\% confidence interval of the approximate filtering distribution.
}
\end{figure}

We address the problem of learning dynamical models that account for epistemic and aleatoric uncertainty.
Our approach allows for epistemic uncertainty by attaching uncertainty to the neural net weights and for aleatoric uncertainty by using a deep state-space formulation (see Sec. \ref{sec:bayesian_dssm}).
While this model family promises flexible predictive distributions, inference is doubly-intractable due to the uncertainty over the weights and the latent dynamics.
The main contribution of this paper is a  
sample-free inference scheme that addresses this pain point and allows us to efficiently propagate uncertainties along a trajectory.
Our deterministic approximation is computationally efficient and accurately captures the first two moments of the predictive distribution.
It can be used as a building block for  multi-step ahead predictions (see Fig. \ref{fig:main_1a}) and Gaussian filtering (see Fig. \ref{fig:main_1b}).
Furthermore, our model approximation can be used as a fully deterministic training objective  (see Sec. \ref{sec:training}).
The runtime of our method is analyzed in Sec. \ref{sec:runtime}.

The paper is complemented by an empirical study (see Sec. \ref{sec:experiments}) that begins with an in-depth examination of each individual building block, showcasing their unique strengths. 
Afterward, we integrate all components and apply our approach to two well-established dynamical modeling benchmark datasets. 
Our method particularly excels in demanding situations, such as those involving noisy transition dynamics or high-dimensional outputs.

\section{Background}
\label{sec:background}
We recap relevant background material before we introduce our model. 
In Sec. \ref{subsec:background_ssm}, we give an introduction to deep state-space models.
Assumed density approximations and Gaussian filtering form the core of our deterministic inference algorithm and are reviewed in Sec. \ref{subsec:background_assumed_density} and Sec. \ref{subsec:background_filtering}.

\subsection{Deep State Space Models}
\label{subsec:background_ssm}

A \textit{State Space Model} (SSM) \cite{saerkkae13_filtering} describes a dynamical system that  is partially observable. The true underlying  process with latent state $x_t\in\mathbb{R}^{D_x}$  emits at each time step $t$ an observation  $y_t\in\mathbb{R}^{D_y}$. The latent dynamics  follow a Markovian structure, i.e., the state at time point $x_{t+1}$ only depends on the state of the previous time point $x_{t}$. 
More formally, the generative model of a SSM can be expressed as
\begin{align}
    x_0 &\sim p(x_0), \label{eq:ssm_init}\\
    x_{t+1} &\sim p(x_{t+1}|x_{t}), \label{eq:ssm_transition}\\
    y_t &\sim p(y_t|x_t).
    \label{eq:ssm_emission}
\end{align}
Above, $p(x_0)$ is the initial distribution, $p(x_{t+1}|x_{t})$ is the transition density, and $p(y_t|x_{t})$ is the emission density.

A \textit{Deep State-Space Model} (DSSM) is a SSM with neural transition and emission densities. Commonly, these densities are modeled as input-dependent Gaussians \cite{krishnan17_dks, bayer2021_mind}. However, there exists also  concurrent work that proposes more expressive densities \cite{bezenac20_nfkalman}.

\subsection{Assumed Density Approximation}
\label{subsec:background_assumed_density}

The $t$-step transition kernel propagates the latent state forward in time and is recursively computed as
\begin{equation}
    p(x_{t+1}|x_0) = \int p(x_{t+1}|x_{t}) p(x_{t}|x_0) d x_{t},
\label{eq:ssm_transition_kernel}
\end{equation}
where $p(x_{t+1} \vert x_{t})$ follows Eq. \eqref{eq:ssm_transition}.
Except for linear transition functions \cite{saerkkae13_filtering}, there
 exists  no analytical solution. 

Various approximations to the transition kernel have been proposed that can be roughly divided into two groups: 
(a) \textit{Monte Carlo} (MC) based approaches \cite{brandt02_simulatedll, petersen95_simulatedll} and
(b) deterministic approximations based on \textit{Assumed Densities} (AD) \cite{saerkkae15_inference}. 
While MC based approaches can, in the limit of infinitely many samples, approximate arbitrarily complex distributions, they are often slow in practice, and their convergence is difficult to assess.
In contrast, deterministic approaches often build on the
assumption that the $t$-step transition kernel can be approximated by a Gaussian distribution.
In the context of machine learning, AD approaches have been recently used in various applications such as deterministic variational inference \cite{wu2018_deterministic} or
traffic forecasting \cite{look2023_cheap}.

We follow the AD approach and approximate the $t$-step transition kernel from Eq. \eqref{eq:ssm_transition_kernel} as 
\begin{align}
p(x_{t+1}|x_0) &\approx \int p(x_{t+1}|x_{t}) \mathcal{N}(x_{t}|m_{t}^x, \Sigma^x_{t} ) d x_{t},\nonumber\\
    &\approx \mathcal{N}(x_{t+1}|m_{t+1}^x, 
    \Sigma^x_{t+1} ).
\label{eq:ssm_transition_kernel_ad}
\end{align}
where the latent state $x_{t}$ is recursively approximated as a Gaussian with mean $m_{t}^x\in \mathbb{R}^{D_x}$ and covariance $\Sigma_{t}^x \in \mathbb{R}^{D_x \times D_x}$. 
This simplifies the calculations for solving Eq. \eqref{eq:ssm_transition_kernel_ad} to approximating the first two output moments.
There exist generic approximation methods 
\cite{solin21_fpk}  as well as specialized algorithms for DSSMs \cite{look2023_cheap}. 
In this work, we will build on the algorithm from \cite{look20_dinsde} that approximates the first two output moments via moment propagation across neural net layers, similarly as \cite{haussmann2021_bayesian, wu2018_deterministic}. 

\subsection{Gaussian Filtering}
\label{subsec:background_filtering}
In filtering applications, we are interested in the distribution $p(x_t|y_{1:t})$, where 
$y_{1:t}=\{y_{1}, \ldots, y_t\}$ denotes the past observations. 
For deep state-space models, the filtering distribution is not tractable, and we can approximate its distribution with a general  Gaussian filter \cite{saerkkae13_filtering, jazwinski1970_filtering} by repeating the subsequent two steps over all time points.
Following concurrent literature \cite{jazwinski1970_filtering}, we refer to 
$p(x_{t}|y_{1:t-1})$ as the prior and 
to $p(x_{t}, y_{t}|y_{1:t-1})$ as the joint prior.

 \textit{Prediction:} 
Approximate the prior $p(x_{t}|y_{1:t-1})$ with
\begin{align}
    p(x_{t}|y_{1:t-1})
    &= \int p(x_{t}|x_{t-1})
    p(x_{t-1}|y_{1:t-1}) d x_{t-1},
    \nonumber
    \\
    &\approx \int p(x_{t}|x_{t-1})
    \mathcal{N}({m}_{t-1}^x, {\Sigma}^x_{t-1} ) d x_{t-1},
    \nonumber
    \\
    &\approx
    \mathcal{N}({m}_{t|t-1}^x, {\Sigma}^x_{t|t-1} ),
    \label{eq:ssm_predict}
\end{align}
where 
$p(x_{t+1}|x_{t})$ refers to the transition model defined in Eq. \eqref{eq:ssm_transition}.
We arrive at Eq. \eqref{eq:ssm_predict} by 
multiple rounds of moment matching. First, we approximate the filtering distribution as a normal distribution, and then we approximate the one-step transition kernel as another normal.
Here, the index $t|t'$ explicitly denotes prior moments, i.e., the moments at time step $t$ conditioned on the observations up to time step $t'$.  If $t=t'$, we omit the double index.

\textit{Update}: Approximate the joint prior $p(x_{t}, y_{t}|y_{1:t-1})$ 
\begin{align}
    p(x_{t}, y_{t}|y_{1:t-1}) 
    &=
    p(y_{t}\vert x_{t}) p(x_{t} \vert y_{1:t-1})  
    \nonumber
    \\
     & \approx
    p(y_{t} \vert x_{t}) \mathcal{N}({m}_{t|t-1}^x, {\Sigma}^x_{t|t-1} ) 
    \nonumber
  \\
    &\approx 
     \mathcal{N}\left(
\begin{bmatrix}
{m}^x_{t|t-1}\\
{m}^y_{t|t-1}\\
\end{bmatrix}
,
\begin{bmatrix}
{\Sigma}_{t|t-1}^{x}  &  {\Sigma}_{t|t-1}^{xy}\\
{\Sigma}_{t|t-1}^{yx} &  {\Sigma}^{y}_{t|t-1}\\
\end{bmatrix}
\right),
 \label{eq:ssm_update2}
\end{align}
where $\Sigma_{t|t-1}^{xy}\in \mathbb{R}^{D_x \times D_y}$ is the cross-covariance between $x_{t}$ and  $y_{t}$ and the density $p(y_{t} \vert x_{t})$ is defined in Eq. \eqref{eq:ssm_emission}.
Building a Gaussian approximation to the joint prior (Eq. \eqref{eq:ssm_update2} can be performed by similar moment matching schemes as discussed in Sec. \ref{subsec:background_assumed_density}.
Afterwards, we can calculate the posterior $p(x_{t}|y_{1:t})$ by conditioning on the observation $y_t$ 
\begin{align}
p(x_{t}|y_{1:t}) &\approx  \mathcal{N}({m}_{t}^x, {\Sigma}^x_{t} ),
 \label{eq:ssm_update3}
\end{align} 
where Eq. \eqref{eq:ssm_update3} can be obtained from Eq. \eqref{eq:ssm_update2} by standard Gaussian conditioning (e.g. \cite{saerkkae13_filtering}).
The resulting distribution has the below moments
\begin{align}
{m}_{t}^x &= m_{t | t-1}^x + K_t(y_t - m_{t | t-1}^y), \label{eq:update_mean} \\
{\Sigma}^x_{t} &= {\Sigma}^x_{t | t-1} - K_t {\Sigma}^y_{t\vert t-1} K_t^\top,
\label{eq:update_cov}
\end{align}
where $K_t \in \mathbb{R}^{D_x \times D_y}$ is the Kalman gain 
\begin{equation}
K_t = \Sigma_{t|t-1}^{xy} \left( \Sigma_{t|t-1}^y \right)^{-1}.
\label{eq:gain}
\end{equation}

Prior work in the context of DSSM and Gaussian Filters \cite{bezenac20_nfkalman} encodes  observations into an auxiliary latent space with an invertible neural net and then relies on a linear SSM formulation in order to be able to exactly solve the Gaussian Filter equations. To the best of our knowledge, there exists no prior work that applies Gaussian Filters on general DSSMs.

\begin{figure}[t]
	\begin{subfigure}[t]{.5\columnwidth}
 \centering
 \includegraphics[height=.8\columnwidth]{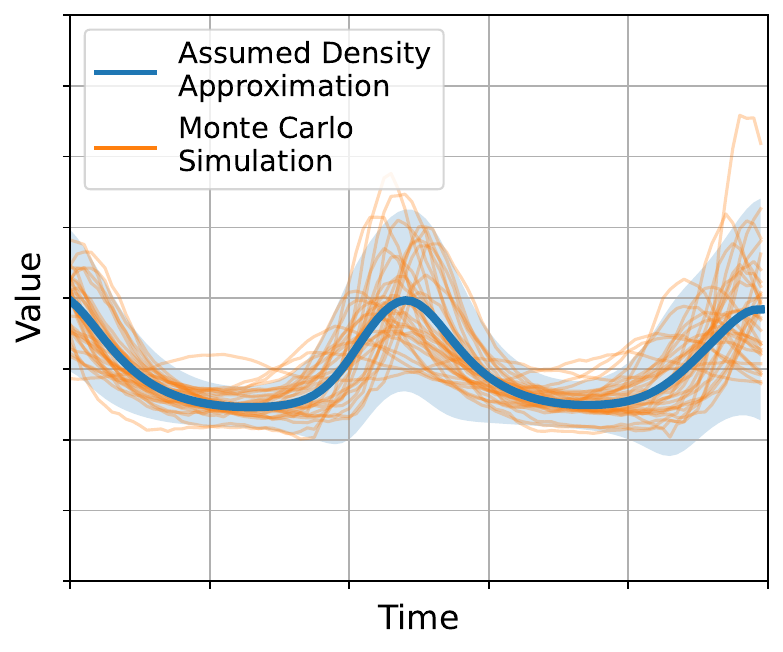}
	\caption{Local Weights.}\label{fig:main_2a}
	\end{subfigure}%
 	\begin{subfigure}[t]{.5\columnwidth}
   \centering
\includegraphics[height=.8\columnwidth]{figures/global.pdf}
	\caption{Global Weights.}\label{fig:main_2b}
	\end{subfigure}%
 \caption{
 Given a dynamical system  $p(x_{t+1}|x_t, w_t)$ with uncertainty over the weights $w_t \sim p(w_t)$  we compare in Panel (a) and (b) two different sampling strategies. In Panel (a) we resample at each time step the weights, while  in Panel (b) the weights are sampled only at the initial time step. We visualize Monte Carlo simulations as orange solid lines and our deterministic output approximation in blue, where the solid line depicts the mean and the shaded area the 95\% confidence interval.}
\end{figure}

\section{Probabilistic  Deep State-Space Models}
\label{sec:bayesian_dssm}

We present our 
 \textit{Probabilistic Deep State-Space Model} (ProDSSM) family in Sec. \ref{subsec:pdssm_uwp}. 
 Our model can account for epistemic uncertainty by attaching uncertainty to the weights of the neural network and for aleatoric uncertainty by building on the deep state-space formalism.
By integrating both sources of  uncertainties, our model family promises well-calibrated uncertainties.
However, joint marginalization over the weights of the neural network and the latent dynamics presents a significant inference challenge.
To this end,
 we present novel algorithms for assumed density approximations (Sec. \ref{subsec:pdssm_adf}) and for Gaussian filtering (Sec. \ref{subsec:pdssm_gf}) that jointly handle the latent states and the weights.
Both algorithms are tailored towards ProDSSMs, allow for fast and sample-free inference with low compute, and lay the basis for our deterministic training objective (Sec. \ref{sec:training}).

\subsection{Uncertainty Weight Propagation}
\label{subsec:pdssm_uwp}

Following \cite{chua18_pets}, we consider two variants of propagating the weight uncertainty along a trajectory: the local and global approach.
For the local approach, we resample the weights $w_t\in \mathbb{R}^{D_w}$ at each time step (see Fig. \ref{fig:main_2a}). 
Contrarily, for 
the global approach, we sample the weights only once at the initial time step and keep them fixed for all remaining time steps (see Fig. \ref{fig:main_2b}).

Assuming Gaussian additive noise, the transition and emission model of ProDSSMs are defined as follows
\begin{align}
x_0&\sim p(x_0),\\
w_0& \sim  p(w_0|\phi),\\
x_{t+1}  & \sim \mathcal{N} \left( x_{t+1} |f(x_{t}, w_{t} ),  \text{diag}(l(x_{t}, w_{t}))  \right), \label{eq:transition} \\
w_{t+1}&\sim
\begin{cases}
    p(w_{t+1}|\phi),   & \text{if {Local}} \\
    {\delta(w_{t+1} - w_0)},& \text{if {Global}}
\end{cases} \label{eq:local_vs_global}\\
y_t &\sim \mathcal{N} \left(y_t|g(x_t),  \text{diag} \left( r \right) \right), \label{eq:emission}
\end{align} 
where 
$f(x_t, w_t): \mathbb{R}^{D_x \times D_w} \rightarrow \mathbb{R}^{D_x}$ models the transition mean, $l(x_t, w_t):\mathbb{R}^{D_x \times D_w} \rightarrow \mathbb{R}_+^{D_x}$ the transition variance,
$g(x_t): \mathbb{R}^{D_x} \rightarrow \mathbb{R}^{D_y}$ the mean emission, and $r \in\mathbb{R}_+^{D_y}$ the emission variance.  
We further model the weight distribution $p(w_t|\phi)$ as a Gaussian distribution  
\begin{equation}
    p(w_t|\phi)=\mathcal{N}(w_t|m^w_t, \text{diag}(\Sigma^w_t)),
    \label{eq:weight_prior}
\end{equation}
with mean $m^w_t\in \mathbb{R}^{D_w}$ and diagonal covariance $\Sigma^w_t \in \mathbb{R}_+^{D_w}$. Both together define the hyperparameters $\phi=\{m^w_t, \Sigma^w_t\}_{t=0}^T$ of our model, where $T$ is the horizon.

In order to avoid cluttered notation, we introduce the augmented state  $z_t= [x_t,  w_t]$ that is a concatenation of the latent state $x_t$ and weight $w_t$, with dimensionality $D_z=D_x + D_w$. 
 The augmented state $z_t$ follows the transition density 
$ \mathcal{N} \left( z_{t+1} |F(z_{t}),  \text{diag}(L(z_{t}))  \right)$, where the mean function $F(z_t): \mathbb{R}^{D_z} \rightarrow \mathbb{R}^{D_z}$ and the covariance function $L(z_t):\mathbb{R}^{D_z} \rightarrow \mathbb{R}_+^{D_z}$ are
 defined as
 \begin{align}
   \left(F(z_{t} ),  L(z_{t} ) \right) = 
    \begin{dcases}
    \left(
    \begin{bmatrix}
    f(x_{t}, w_t)\\
   m_{t+1}^w
    \end{bmatrix},
    \begin{bmatrix}
    l(x_{t}, w_t)\\
     \Sigma^w_{t+1}
    \end{bmatrix}
    \right)
    & \text{if {Local}}  \\
   \left(
   \begin{bmatrix}
    f(x_{t}, w_t)\\
   w_t
    \end{bmatrix},
     \begin{bmatrix}
    l(x_{t}, w_t) \\
    0
    \end{bmatrix}
    \right)
    & \text{if {Global}}.
\end{dcases}
\label{eq:augmented_state}
\end{align}

In the following, we extend the moment matching algorithm from \cite{look20_dinsde} towards  ProDSSMs and Gaussian filters.  
Our algorithmic advances are general and can be combined with both weight uncertainties propagation schemes.

\subsection{Assumed Density Approximation}
\label{subsec:pdssm_adf}
In this section, we present our main contribution, which is a novel approximation to the $t$-step transition kernel $p(z_{t+1}|z_0)$ for ProDSSMs. 
Our approximation takes an assumed density approach and propagates moments along time direction and across neural network layers, similarly, as in \cite{look20_dinsde}. 
Prior work either deals with non-recurrent neural network architectures \cite{wu2018_deterministic} or deterministic weights \cite{look20_dinsde}, while our new model family, ProDSSM, requires both.
In contrast to prior work, we need to account for the correlation between weights and states. 

We follow the general assumed density approach (see Sec. \ref{subsec:background_filtering}) on the augmented state $z_t$.
As a result, we obtain a Gaussian approximation 
$p(z_{t+1}| z_0) \approx  \mathcal{N}(z_{t+1}|m_{t+1}^z, \Sigma_{t+1}^z)$ to the $t$-step transition kernel that approximates the joint density over the latent state $x_t$ and the weights $w_t$.
The mean and the covariance have the structure
\begin{align}
    m_{t}^z &=
  \begin{bmatrix}
  m_t^x\\
  m_t^w\\
   \end{bmatrix},
   &
   \Sigma_t^z = 
     \begin{bmatrix}
   \Sigma_{t}^{x}  &  \Sigma_t^{xw} \\
    \Sigma_t^{wx} &  \Sigma_t^{w} \\
   \end{bmatrix}
\end{align}
where $\Sigma_t^{x}\in \mathbb{R}^{D_x \times D_x}$ is the covariance  of $x_t$ and
$\Sigma_t^{xw}\in \mathbb{R}^{D_x \times D_w}$ is the cross-covariance between $x_t$ and  $w_t$. 

For a standard DSSM architecture, the number of weights exceeds the number of latent dimensions. 
Since the mean and the covariance over the weights are not updated over time,
the computational burden of computing $\Sigma_t^z$ is dominated by the computation of the cross-covariance $\Sigma_t^{xw}$.
This covariance becomes zero for the local approach due to the resampling step at each time point.
Consequently, the local approach exhibits reduced runtime and memory complexity compared to the global approach.

In the following, we will detail how the remaining terms can be efficiently computed by propagating moments through the layers of a neural network.
We start by applying the law of unconscious statistician, which tells us that the moments of the augmented state at time step $t+1$ are available as a function of prior moments at time step $t$ \cite{look20_dinsde}
\begin{align*}
    m_{t+1}^z = \mathbb{E}[F(z_t)],
    &&
    \Sigma_{t+1}^z = \text{Cov}[F(z_t)] + \text{diag}(\mathbb{E}[L(z_t)]).
    \numberthis
    \label{eq:prior_augmentedstate}
\end{align*}
Now, we are left with calculating the first two output moments of the augmented mean $F(z_t)$ and covariance update $L(z_t)$. 
In the following, we discuss the approximation of the output moments for the augmented $F(z_t)$ and omit the discussion on the augmented covariance update  $L(z_t)$ as its moments can be approximated similarly.
Typically, neural networks are a composition of $L$ simple functions (layers)  that allows us to write the output as 
$F(z_t)= U^L(\ldots U^1(z^0_t)\ldots), $
where $z_t^l \in \mathbb{R}^{D_z^l}$ is the augmented state at layer $l$ at time point $t$. We denote the input as $z_t^0=z_t$.
The function  $U^{l}(z_t^{l-1}): \mathbb{R}^{D_z^{l-1}} \rightarrow \mathbb{R}^{D_z^{l}}$ at the $l$-th layer  receives the augmented state $z^{l-1}_t$ from the previous layer and calculates the output $z^l_t$  as
\begin{align}
    U^{l}(z^{l-1}_t)&= 
     \begin{bmatrix}
      x^{l}_t\\
      w^l_t
    \end{bmatrix}
    =
    \begin{bmatrix}
      u^{l}(x^{l-1}_t, w_t^{l-1})\\
      w^{l-1}_t
    \end{bmatrix},
\end{align} 
where $x_t^{l} \in \mathbb{R}^{D_x^l}$ is the state at layer $l$ at time point $t$ and  $u^{l}(x_t^{l-1}, w_t^{l-1}): \mathbb{R}^{D_x^{l-1}} \times \mathbb{R}^{D_w} \rightarrow \mathbb{R}^{D_x^{l}}$ is the function that updates the state. The weights $w_t^{l} \in \mathbb{R}^{D_w}$ 
are not altered in the intermediate layers and the last layer returns the weight for the global approach or its mean $m_t^w$ for the local approach.
We approximate the output distribution of each layer recursively  as
\begin{equation}
    p(z^l_t)=p(U^l(z^{l-1}_t))
    \approx  \mathcal{N}(z^l_t|m^l_t, \Sigma^l_t),
    \label{eq:layerwise_moment_propagation}
\end{equation}
where $m^l_t \in \mathbb{R}^{D_z^{l}}$ and $\Sigma^l_t\in \mathbb{R}^{D_z^{l}\times D_z^l}$ are the mean and covariance of ${z}^l_t$. 
We refer to calculating $m^l_t$ and $\Sigma^l_t$ for each layer as layerwise moment propagation \cite{look20_dinsde}. 
In the remainder of this subsection, we will present the output moments for the linear layer and ReLU activation function for the global as well as local approach. 

\subsubsection{Output Moments of the Linear Layer}
A linear layer applies an affine transformation 
\begin{align}
     U({z}^l_t) =  
     \begin{bmatrix}
    A^l_t {x}_t^l + b^l_t\\
     w_t^l
     \end{bmatrix},  
 \end{align}
 where the transformation matrix $A^l_t\in \mathbb{R}^{D_x^{l+1} \times D_x^{l}}$ and bias $b^l_t\in \mathbb{R}^{D_x^{l+1}}$ are both part of weights $(A^l_t,b^l_t) \in w_t^l$. We note that the set of all transformation matrices and biases $\{(A_t^l, b_t^l) \}_{l=1}^L$ define the weights $w_t^l$. 
 As the cross-covariance matrix $\Sigma_t^{l,xw}$ is non-zero  for global weights, the  transformation matrix $A^l_t$, bias $b^l_t$, and state $x_t^l$ are assumed to be jointly normally distributed. 

The mean and the covariance of the weights $w_t$ are equal to the input moments due to the identity function.
The remaining output moments of the affine transformation can be calculated as
\begin{align}
    m^{l+1,x}_{t} &= 
    \mathbb{E}[A^l_t {x}^l_{t}] + \mathbb{E}[b_t^l], \label{eq:affine_mean_x_main}\\
     \Sigma^{l+1, x}_{t}
    &=
    \text{Cov}[A^l_t {x}^l_{t}, A^l_t {x}^l] 
    +    \text{Cov}[b^l_t, A^l_t {x}^l_{t}] 
    \nonumber \\
      &
     ~~~~+    \text{Cov}[ A^l_t {x}^l_{t}, b^l_t]
     + \text{Cov} [b^l_t, b^l_t],
     \label{eq:affine_cov_xx_main}
    \\
    \Sigma^{l+1, xw}_{t}&=
    \text{Cov}[A^l_t {x}^l_{t}, w^l] 
    +
    \text{Cov}[b^l_t, w_t^l], \label{eq:affine_cov_xw_main}
\end{align}
which is a direct result of the linearity of the $\text{Cov}[\bullet, \bullet]$ operator.
In order to compute the above  moments, we need to  calculate the 
moments of a product of correlated normal variables, $\mathbb{E}[A^l_t{x}_{t}^l], \text{Cov}[
    A^l_t{x}_{t}^l,
    A^l_t{x}_{t}^l],$ and $\text{Cov}[
    A^l_t{x}_{t}^l,w^l]$.
Surprisingly, these computations can be performed in closed form for both local and global weights provided that $x_t^l$ and $w_t^l$ follow a normal distribution.   
We provide a detailed derivation and the final results in App. \ref{app:moments_linear_layer}.

For the case of local weights, the cross-covariance matrix $\Sigma_t^{l,xw}$ becomes zero, i.e., weights and states are uncorrelated.
In addition, the computation of the remaining terms simplifies significantly (see also App. \ref{app:moments_linear_layer}), and, as a result, we can recover the results from \cite{wu2018_deterministic}.

\subsubsection{Output Moments of the ReLU Activation}

The ReLU activation function applies element-wise the $\text{max}$-operator to the latent states while the weights stay unaffected
\begin{align}
    U({z}_t^l)= \begin{bmatrix}
    \text{max}(0, {x}^{l}_t)\\
    w_t^l
    \end{bmatrix}.
\end{align}
Mean $m_t^{l+1,x}$ and covariance $\Sigma_t^{l+1, x}$ of the state $x_t^{l+1}$ are available in related literature \cite{wu2018_deterministic}.
Mean $m_t^{l+1,w}$ and covariance $\Sigma_t^{l+1, w}$ of the state $w_t^{l+1}$ are equal to the input moments, $m_t^{l,w}$ and $\Sigma_t^{l, w}$.
For the case of global weights, it remains open to calculate the cross-covariance $\Sigma_t^{l+1,xw}$. 

Using Stein's lemma \cite{liu1994_steinslemam}, we can calculate the cross-covariance after the ReLU activation as
\begin{equation}
    \Sigma_t^{l+1,xw} =
    \mathbb{E}[\nabla_{{x}^l_t} \text{max}(0, {x}^{l}_t) ] \Sigma_t^{l, xw},
\end{equation}
where $\mathbb{E}[\nabla_{{x}^l_t} \text{max}(0, {x}^{l}_t) ]$ is the expected Jacobian of the ReLU activation.
The expected Jacobian is equal to the expectation of the Heaviside function, which can be closely approximated \cite{wu2018_deterministic}.

\subsection{Gaussian Filtering}
\label{subsec:pdssm_gf}
Our approximation to the filtering distribution, $p(z_t|y_{1:t})$, follows the Gaussian filter (see Sec. \ref{subsec:background_filtering}).
In contrast to prior work, we extend the filtering step to the augmented state consisting of the latent dynamics and the weights.
In standard architectures, the number of latent states is small compared to the number of weights, which makes filtering in our new scenario more demanding.
We address this challenge by applying our deterministic moment matching scheme that propagates moments across neural network layers.
Additionally, we combine it with our previously derived approximation to the $t$-step transition kernel $p(z_{t+1}|z_0)$ from Sec. \ref{subsec:pdssm_adf}.
We also verify empirically in Sec. \ref{subsec:experiments_filtering} that standard numerical integration schemes are not well suited for filtering tasks of this type.

The Gaussian filter alternates between the prediction and the update step. 
In the following, we explain in more detail how our deterministic moment matching scheme can be integrated into both steps.
For the prediction step, Eq. \eqref{eq:ssm_predict},
we can reuse the assumed density approach that we just derived in order to compute a Gaussian approximation to the predictive distribution $p(z_t \vert y_{1:t-1})$.

For the update step, we need to first find a Gaussian approximation to the joint distribution of the augmented state $z_t$ and observation $y_t$ conditioned on $y_{1:t-1}$ (see also Eq. \eqref{eq:ssm_update2})
\begin{equation}
    p(z_t, y_t | y_{1:t-1}) \approx 
    \mathcal{N}\left(
    \begin{bmatrix}
  m_{t|t-1}^z\\
  m_{t|t-1}^y\\
   \end{bmatrix}
   ,
   \begin{bmatrix}
   \Sigma_{t|t-1}^{z}  &  \Sigma_{t|t-1}^{zy}\\
   \Sigma_{t|t-1}^{yz} &  \Sigma^{y}_{t|t-1}\\
   \end{bmatrix}
    \right).
\end{equation}
The mean and the covariance of the latent state $z_t$ are known from the prediction step, while their equivalents of the emission $y_t$ are available as
\begin{align}
    m_{{t|t-1}}^y = \mathbb{E}[g(x_t)],
    &&
    \Sigma_{{t|t-1}}^y = \text{Cov}[g(x_t)] + \text{diag}(r),
    \label{eq:prior_observation}
\end{align}
with $x_t \sim \mathcal{N}(m^x_{t|t-1}, \Sigma_{t|t-1}^x)$.
These moments can be approximated  with layerwise moment propagation, as described in the previous section. 
Finally, we facilitate the computation of the cross-covariance $\Sigma_{t|t-1}^{yz}$ be using Stein`s lemma \cite{liu1994_steinslemam}
\begin{align}
    \Sigma_{t|t-1}^{yz}=  
    \text{Cov}[g(x_t), z_t] 
    =
    \mathbb{E}[\nabla_{{x_t}} 
    g(x_t)] \Sigma_{t|t-1}^{xz}.
    \label{eq:prior_crosscovariance}
\end{align}
where the expected Jacobian $\mathbb{E}[\nabla_{{x_t}}g(x_t)]$ of the mean emission function cannot be computed analytically.
We follow the approximation of \cite{look20_dinsde} that reduces the computation to estimate the expected Jacobian per layer.
The latter is often available in closed form, or close approximations exist.

Once we have calculated the joint distribution, we approximate the conditional as another normal distribution, $p(z_t|y_{1:t}) \approx   \mathcal{N}(m_t^z, \Sigma_t^z)$, 
as shown in Eq. \eqref{eq:gain}.
For the global approach, the Kalman gain has the structure 
$K_t = \Sigma_t^{zy}(\Sigma_t^y)^{-1}$, and 
the updated covariance matrix $\Sigma_t^z$ of augmented state $z_t$ is dense. As a consequence, the weights $w_t$ have a non-zero correlation after the update, and the overall variance gets reduced.
For the local approach, only the distribution of the states $x_t$ will be updated since the lower block of the gain matrix is zero. 
The weight distribution, as well as the cross-covariance between the states and weights, is hence not affected by the Kalman step.
 \section{Training and Predictions}
\label{sec:training}

In this section, we derive efficient and sample-free training and testing routines for ProDSSMs.
These routines build on the assumed density approximation and the Gaussian filter that we introduced in the previous section.

\subsection{Training}
\label{subsec:trainig}

We train the ProDSSMs by fitting
the hyperparameters $\phi$ to a dataset $\mathcal{D}$. 
The hyperparameters $\phi$ describe the weight distribution. 
For the sake of brevity, we introduce the shorthand notation $p(w_{0:T}|\phi)=p(w|\phi)$ to refer to the weights at all time steps with arbitrary horizon $T$.
We propose to train the ProDSSM on a Type-II \textit{Maximum A Posteriori} (MAP) objective 
(see \cite{murphy2013_machine} Chap. 5.6) 
\begin{equation}
    \underset{\phi}{\text{argmax}}
    \log \int p(\mathcal{D}|w) p(w|\phi) dw
    + \log p(\phi).  \label{eq:type2_map}
\end{equation}
This objective is also termed as predictive variational
Bayesian inference by \cite{futami22_pbvi} as it directly minimizes the  Kullback-Leibler divergence between the true data generating distribution and the predictive
distribution, which we aim to learn.  
Compared to other learning objectives, Eq. \eqref{eq:type2_map} provides better predictive performance, is more robust to model misspecification, and provides a beneficial implicit regularization effect for over-parameterized models. We refer to \cite{masegosa20_modelmiss, morningstar22a_pacm, futami22_pbvi} that studies this learning objective for probabilistic neural nets in more detail from a theoretical as well as an empirical point of view.

In our work, we show that the typically hard to evaluate likelihood  
$p(\mathcal{D}|\phi)=\int  p(D\vert w)p(w \vert \phi) dw$
can be closely approximated with deterministic moment matching routines. 
The exact form of the likelihood  hereby depends on the task at hand, and we specify in our experiments how the likelihood can be closely approximated for regression problems in Sec. \ref{subsec:experiments_continuousdepthlayers} and for dynamical system modeling in Sec. \ref{subsec:experiments_dynamicalsystems}.

We are now left with defining the hyper-prior $p(\phi)$. Remember, $\phi$ defines the  weight distribution that is defined by its two first moments $m^w=m^w_{0:T}$ and $\Sigma^w=\Sigma_{0:T}^w$.
In order to arrive at an analytical objective, we model each entry in $p(\phi)$ independently. We define the hyper-prior of the $i$-th entry of the mean as a standard Normal 
\begin{align}
    \log p(m^w_{i}) &= \log \mathcal{N}(m^w_i|0, I) \nonumber\\
    &= -\frac{1}{2} (m_{i}^w)^2 + \text{const.}
\end{align}
and, assuming that the covariance is diagonal, chose the Gamma distribution for the $(i,i)$-th covariance entry
\begin{align}
    \log p(\Sigma_{ii}^w) &=\log \text{Ga}(\Sigma^w_{ii}|\alpha=1.5, \beta=0.5)\nonumber\\
    &= \frac{1}{2}\log \Sigma_{ii}^w - \frac{1}{2} \Sigma_{ii}^w
    + \text{const.},
\end{align}
where $\alpha$ is the shape parameter and $\beta$ is the rate parameter. We insert the above hyper-prior of the mean and covariance into $\log p(\phi)$ and arrive at
\begin{align}
   \log p(\phi) &= \log p(m^w) + \log p(\Sigma^w) \nonumber\\
   &=\frac{1}{2}\sum_{i=1}^{D_w}
     \log \Sigma_{ii}^w -\! (m_{i}^w)^2 -\! \Sigma_{ii}^w 
     +\!\text{const.},
     \label{eq:hyper-prior}
\end{align}
which leads to a total of $2 D_w$ hyperparameters, i.e., one for the mean and one for the variance of each weight.

In contrast, the classical Bayesian formalism keeps the prior $p(w \vert \phi)$ constant during learning and 
the posterior $p(w|\mathcal{D})$ is the quantity of interest.
As an analytical solution to the posterior is intractable, either \textit{Markov Chain Monte Carlo} (MCMC) \cite{maddox19_swag} or \textit{Variational Inference} (VI) \cite{graves11_practicalvi} is used.
It is interesting to note that the only difference between our formulation and the objective in VI, with a suitable prior choice, is the position of 
the logarithm in the likelihood  $p(\mathcal{D}|\phi)$. 
Please see App. \ref{app:background_bnn} for more details. 
However, we are not aware of any prior work that applies VI in the context of ProDSSMs. 
Closest to our work is most likely \cite{depeweg2017_decomposition} that approximates the posterior over the weights for fully observed stochastic dynamical systems, i.e., without latent states. 

\subsection{Predictive Distribution}
\label{subsec:predictive_distribution}

 During test time, we are interested in the predictive distribution $p(y_{t}|y_{-H:0})$ at   time step $t$ conditioned on the observations $y_{-H:0}=\{y_{-H}.\ldots, y_0\}$ with  conditioning horizon $H\in \mathbb{N}_+$. 
 The predictive distribution is computed as
 \begin{align}
     p(y_{t}| y_{-H:0}) &=
\int p(y_{t}|z_{t})p(z_{t}|z_0)p(z_0|y_{-H:0}) d z_0, z_{t},
\nonumber
\\
     &= \int p(y_{t}|z_{t}) p(z_t|y_{-H:0}) d z_{t}.
     \label{eq:predictive_distribution}
 \end{align}
 Above,  $p(z_0|y_{-H:0})$ is the filtering distribution,  $p(z_{t}|z_0)$ is the $t$-step transition kernel and $p(z_t \vert y_{-H:0})$ the $t$-step marginal.
 Prior work on general deep SSMs  \cite{bayer2021_mind, chung15_vrnn, krishnan17_dks} relies on auxiliary networks in order to approximate the filtering distribution and then uses MC integration in order to compute predictive distribution. 
Contrarily, we replace the need for auxiliary networks and MC integration with our deterministic moment matching scheme.

The computation of the predictive distribution is performed by 
 a series of Gaussian approximations:
\begin{align}
     p(y_{t}| y_{-H:0}) &\approx     \int p(y_{t}|z_{t})p(z_{t}|z_0)
     \mathcal{N}(m_0^z, \Sigma_0^z)dz_0, z_t
     \nonumber \\
     &\approx     \int p(y_{t}|z_{t})
     \mathcal{N}(m_{t|0}^z, \Sigma_{t|0}^z)d z_t
     \nonumber\\
      &\approx     
     \mathcal{N}(m_{t|0}^y, \Sigma_{t|0}^y),
     \label{eq:predictive_distribution_AD}
\end{align}
where the density $ \mathcal{N}(m_0^z, \Sigma_0^z)$  approximates the filtering distribution. 
Its computation is described in Sec. \ref{subsec:pdssm_gf}.
We obtain the density $ \mathcal{N}(m_{t|0}^z, \Sigma_{t|0}^z)$ as an approximation to the $t$-step marginal kernel $ p(z_t|y_{-H:0})$ in Eq. \eqref{eq:predictive_distribution} by propagating the augmented latent state forward in time as described in Sec. \ref{subsec:pdssm_adf}.
Finally, we approximate the predictive distribution $p(y_{t}| y_{-H:0})$ with the density $\mathcal{N}(m_{t|0}^y, \Sigma_{t|0}^y)$ in Eq. \eqref{eq:predictive_distribution_AD},
which can be done by another round of moment matching as also outlined in Eq. \eqref{eq:prior_observation}.

 We present pseudo-code for approximating the predictive distribution in Alg. \ref{alg:predictive} that relies on Alg. \ref{alg:filtering} to approximate the filtering distribution $p(z_0|y_{-H:0})\approx \mathcal{N}(z_0|m_0^z,\Sigma_0^z)$ 
 Both algorithms explicitly do a resampling step for the local weight setting. In practice, it is not necessary, and we just omit the calculation.

\begin{algorithm}
    \scriptsize
	\caption{Deterministic Inference \texttt{DetInf}}\label{alg:predictive}
	\begin{algorithmic}
 \State {\bf Inputs:}  $f(x_t, w_t)$ \Comment{Mean update}
	\State ~~~~~~~~~~~  $l(x_t, w_t)$  \Comment{Covariance update}
	\State ~~~~~~~~~~~  $g(x_t)$  \Comment{Mean emission}
	\State ~~~~~~~~~~~  $r$  \Comment{Covariance emission}
	\State ~~~~~~~~~~~  $p(z_{-H})$  
 \Comment{Initial distribution}
 \State ~~~~~~~~~~~  $y_{-H:0}$  \Comment{Observations}
 
 \State {\bf Outputs:}  $p(y_T|y_{-H:0}) \approx \mathcal{N}
 (y_{T}|m^y_{T|0}, \Sigma_{T|0}^y)$ \Comment{Predictive Distribution}
 \State
$m_0^z, \Sigma_0^z \leftarrow $ \texttt{DetFilt}$(f, l, g, r, p(z_{-H}), y_{-H:0})$
 \For{ time step $t \in \{0,\cdots,T-1\}$ }
 \If{Local}
 \State $m_{t|0}^{w}, \Sigma_{t|0}^{w}, \Sigma_{t|0}^{xw},\Sigma_{t|0}^{wx} \leftarrow m_{-H}^{w}, \Sigma_{-H}^{w}, 0,0$
 \Comment{Resample}
 \EndIf
\State
$m_{t+1|0}^z \leftarrow \mathbb{E}[F(z_{t})]$
\Comment{Eq. \ref{eq:prior_augmentedstate}}
\State
$\Sigma_{t+1|0}^z \leftarrow \text{Cov}[F(z_{t})] + \text{diag}(\mathbb{E}[L(z_{t})])$
\Comment{Eq. \ref{eq:prior_augmentedstate}}
\State
$p(z_{t+1}|y_{-H:0}) \leftarrow \mathcal{N}
(z_{t+1}|m_{t+1|0}^z, \Sigma_{t+1|0}^z)$
  \EndFor
\State
$m_{T|0}^y \leftarrow \mathbb{E}[g(x_{T})]$
\Comment{Eq. \ref{eq:prior_observation}}
\State
$\Sigma_{T|0}^y \leftarrow \text{Cov}[g(x_{T})] + \text{diag}(r)$
\Comment{Eq. \ref{eq:prior_observation}}
 \State \Return $\mathcal{N}
 (y_{T}|m^y_{T|0}, \Sigma_{T|0}^y)$
\end{algorithmic}
\end{algorithm}

\begin{algorithm}
    \scriptsize
	\caption{Deterministic Filtering (\texttt{DetFilt}) }\label{alg:filtering}
	\begin{algorithmic}
 \State {\bf Inputs:}  $f(x_t, w_t)$ \Comment{Mean update}
	\State ~~~~~~~~~~~  $l(x_t, w_t)$  \Comment{Covariance update}
	\State ~~~~~~~~~~~  $g(x_t)$  \Comment{Mean emission}
	\State ~~~~~~~~~~~  $r$  \Comment{Covariance emission}
	\State ~~~~~~~~~~~  $p(z_{0})$  
 \Comment{Initial distribution}
 \State ~~~~~~~~~~~  $y_{1:T}$  \Comment{Observations}
 
 \State {\bf Outputs:}  $p(z_T|y_{1:T}) \approx \mathcal{N}
 (z_{T}|m^z_{T}, \Sigma_{T}^z)$ \Comment{Filtering Distribution}
 \State
$p(z_{0}|y_{1:0}) \leftarrow p(z_{0})$
 \For{ time step $t \in \{0,\cdots,T-1\}$ } 
 \If{Local}
 \State $m_{t}^{w}, \Sigma_{t}^{w}, \Sigma_{t}^{xw},\Sigma_{t}^{wx} \leftarrow m_{0}^{w}, \Sigma_{0}^{w}, 0,0$
 \Comment{Resample}
 \EndIf
\State
$m_{t+1|t}^z \leftarrow \mathbb{E}[F(z_{t})]$
\Comment{Eq. \ref{eq:prior_augmentedstate}}
\State
$\Sigma_{t+1|t}^z \leftarrow \text{Cov}[F(z_{t})] + \text{diag}(\mathbb{E}[L(z_{t})])$
\Comment{Eq. \ref{eq:prior_augmentedstate}}
\State
$m_{t+1|t}^y \leftarrow \mathbb{E}[g(x_{t})]$
\Comment{Eq. \ref{eq:prior_observation}}
\State
$\Sigma_{t+1|t}^y \leftarrow \text{Cov}[g(x_{t})] + \text{diag}(r)$
\Comment{Eq. \ref{eq:prior_observation}}
 \State
$\Sigma_{t+1|t}^{yz} \leftarrow \mathbb{E}[\nabla_{{x_{t+1}}} 
    g(x_{t+1})] \Sigma_{t+1|t}^{xz}$
\Comment{Eq. \ref{eq:prior_crosscovariance}}
\State
$K_{t+1} \leftarrow \Sigma_{t+1|t}^{zy}(\Sigma_{t+1|t}^y)^{-1}$
\Comment{Eq. \ref{eq:gain}}
 \State
$m_{t+1}^{z} \leftarrow 
m_{t+1|t}^z + K_{t+1}(y_{t+1}-m_{t+1|t}^y)$
\Comment{Eq. \ref{eq:update_mean}}
  \State
$\Sigma_{t+1}^{z} \leftarrow 
\Sigma_{t+1|t}^z - K_{t+1}\Sigma_{t+1|t}^y K_{t+1}^T$
\Comment{Eq. \ref{eq:update_cov}}
 \State
$p(z_{t+1}|y_{1:t+1}) \leftarrow \mathcal{N}(z_{t+1}|m_{t+1}^z, \Sigma_{t+1}^z)$
 \EndFor
    \State \Return $\mathcal{N}
 (z_{T}|m^z_{T}, \Sigma_{T}^z)$
\end{algorithmic}
\end{algorithm}

\section{Runtime}
\label{sec:runtime}
We first analyze the theoretical runtime of our algorithm  in Sec. \ref{subsec:runtime_theoretical} and then measure its wall clock time in Sec. \ref{subsec:runtime_measured}. 

\subsection{Theoretical Runtime}
\label{subsec:runtime_theoretical}
In our theoretical runtime analysis, we first investigate the runtime for simulating forwards in time and, secondly, the runtime for filtering applications.
We further assume that we have a ProDSSM with maximal hidden layer width $H$ and that the dimensions of $D_x$ and $D_y$ are less than or equal to $H$.

Independent of the weight modeling scheme, 
predicting the next observation $x_{t+1}$ conditioned on the latent state $x_t$ is done by propagating the state through a series of affine transformations and non-linear activities.
The affine transformations scale polynomially with the hidden layer width, whereas the non-linearities are elementwise operations and can be neglected.

Approximating the first two output moments (see. Eq. \ref{eq:prior_observation}) by MC simulation requires propagating $S$ particles, resulting thus in a total cost of $\mathcal{O}(SH^2)$.
Our method approximates the $S\rightarrow \infty$ limit.
For global weights, the computational cost of our method is $\mathcal{O}(H^4 + D_w H^2)$ where $D_w$ is the number of weight parameters.
The first term, $\mathcal{O}(H^4)$, is due to the computational cost of the covariance $\text{Cov}[A^l_t x_t^l, A^l_t, x_t^l] \in \mathbb{R}^{H \times H}$ in Eq. \eqref{eq:affine_cov_xx_main}, where the computation of each matrix entry scales with $O(H^2)$ due to the linearity of the covariance operator.
The second term, $\mathcal{O}(H^2 D_w)$, is due to the cross-covariance $\text{Cov}[A^l_t x_t^l, w^l_t] \in \mathbb{R}^{H \times W}$ in Eq. \eqref{eq:affine_cov_xw_main}, where the computation of each entry scales with $O(H)$, again due to the linearity of the covariance operator.
For local weights, the weights and the states are independent. 
As a result, we can simplify the computation of the first term to $\text{Cov}[A^l_t x_t^l, A^l_t, x_t^l]= \mathbb{E}[A^l_t] \text{Cov}[x_t^l, x_t^l] \mathbb{E}[A^l_t]^{\top}$ and the second term, $\text{Cov}[A^l_t x_t^l, w^l_t]$ becomes zero.
This leads to a runtime reduction to $O(H^3)$.


Our filtering algorithm  necessitates
$\mathcal{O}(H^3)$ 
computations to approximate the output moments of the emission independent of the weight modeling scheme. 
For global weights approximating the cross-covariance between the emissions and augmented latent state involves 
$\mathcal{O}(H^3+H^2D_w)$ 
computations.
Forming the gain matrix involves 
$\mathcal{O}(H^3 + H^2 D_w)$
computations. The first term is caused by inverting the covariance matrix of the emissions, and the second term  is caused by multiplying the inverse covariance matrix with the cross-covariance of the augmented latent state 
(see Eq. \ref{eq:gain}). 
Lastly, updating the moments of latent state (see Eq. \ref{eq:update_cov}) involves 
$\mathcal{O}(H(H+D_w)^2)$
computations which is the most time-consuming step and dominates the total runtime. 
 Similarly, the computational cost of our algorithm for local weights can be derived and has a total cost of $\mathcal{O}(H^3)$.

\subsection{Measured Runtime}
\label{subsec:runtime_measured}
\begin{figure}[t]
\centering
\begin{subfigure}[t]{.48\columnwidth}
   \centering
	\includegraphics[height=.7\columnwidth]{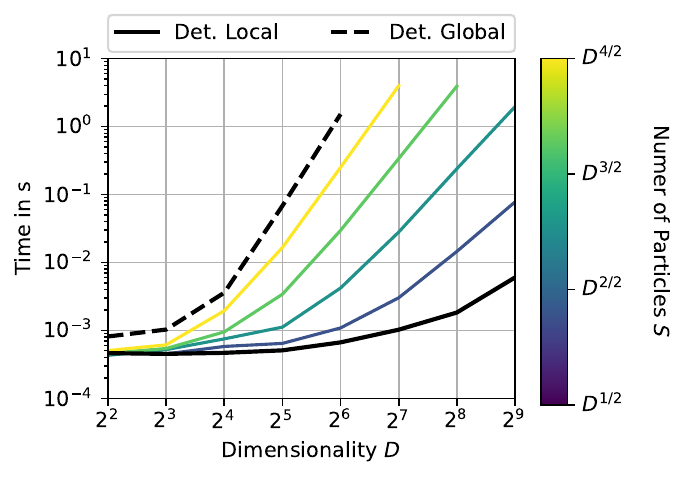}
	\caption{MC runtime including sampling.}\label{fig:time_w_sampling}
	\end{subfigure}%
 \hfill
	\begin{subfigure}[t]{.48\columnwidth}
 \centering
	\includegraphics[height=.7\columnwidth]{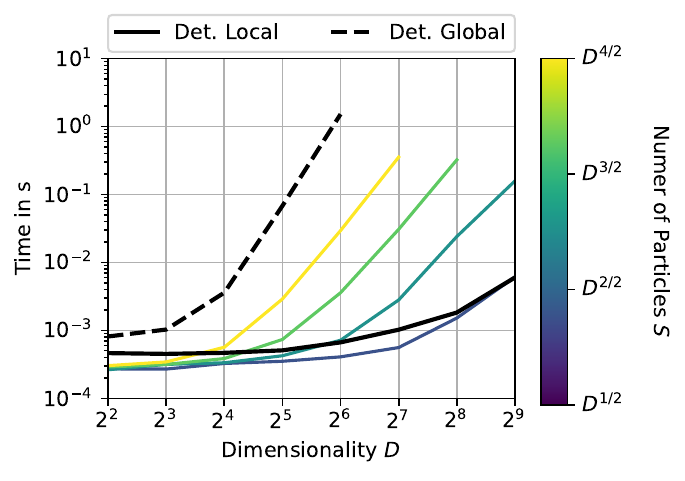}
	\caption{MC runtime measured without sampling.}\label{fig:time_wo_sampling}
	\end{subfigure}
 \caption{
 We visualize the runtime of approximating mean $m_{t+1}^y$ and covariance $\Sigma_{t+1}^y$ of the observation $y_{t+1}$ conditioned on the augmented state $z_t$ at the prior time step with mean $m_{t}^z$ and covariance $\Sigma_{t}^z$. 
 We vary on the x-axis the dimensionality $D$. We use the same dimensionality for the observation $y_t$ and latent state $x_t$, i.e., $D_x=D_y=D$.
 We use randomly initialized transition and emission functions with one hidden layer of width $H=D$.
 The solid/dashed line represents the runtime of our deterministic approximation for local/global weights. The  colored lines represent the runtime of the MC approximation with varying number of particles $S$ as a function of dimensionality $D$. In the left panel, we take into account the runtime of the weight sampling procedure for the MC baseline. In the right panel, we ignore the runtime of the weight sampling procedure. }
 \label{fig:time}
\end{figure}

We visualize in Fig. \ref{fig:time} the wallclock time of our method for approximating the mean and covariance of the observation $y_{t+1}$ conditioned on the mean and covariance of the latent state $x_t$ at the prior time step.
Additionally, we visualize the runtime of the MC baseline with different sampling strategies as a function of the dimensionality $D=D_x=D_y=H$. 
The early stops indicate when we run out of memory. 
We conduct the experiment on a CPU with 32GB memory.
For $S=D$ particles the MC baseline has the same theoretical runtime as our method for local weights. 
In practice, we observe our method for local weights to be faster than the MC baseline with $S=D$ when we include the runtime of the weight sampling procedure. When we exclude the runtime of the weight sampling procedure, our method is faster for $D>64$.
Furthermore, our method for global weights is slower and runs out of memory earlier than all baselines. 
We leave optimizing the runtime of our method for global weights 
as a direction for future work. 
 \section{Experiments}
\label{sec:experiments}
Our paper provides an efficient and sample-free algorithm for learning unknown dynamics from data. By taking epistemic and aleatoric uncertainty into account, our model family, ProDSSM, can produce flexible and well-calibrated predictions over a wide range of scenarios.
Core to our algorithm is a new moment matching scheme that can be applied for assumed density approximation (see Sec. \ref{subsec:pdssm_adf}) and for Gaussian filtering (see Sec. \ref{subsec:pdssm_gf}).
In our experiments, we first analyze each of these algorithmic advances in isolation before putting everything together. 

For this, we first explore in Sec. \ref{subsec:experiments_continuousdepthlayers} our assumed density approximation in the context of deep stochastic layers on eight UCI datasets.  
Then, we study our approximation to the Gaussian Filter in Sec. \ref{subsec:experiments_filtering} on a non-linear filtering task.
We connect both steps and benchmark our full method in Sec. \ref{subsec:experiments_dynamicalsystems} on two well-established dynamical modeling datasets.  
Finally, we summarize our empirical findings in Sec. \ref{subsec:experiments_summary}.

\subsection{Deep Stochastic Layers}
\label{subsec:experiments_continuousdepthlayers}
We first demonstrate the usefulness of our uncertainty propagation scheme as proposed in Sec. \ref{subsec:pdssm_adf} on a regression task with inputs $x \in \mathbb{R}^{D_x}$ and outputs $y \in \mathbb{R}$. 
Here, we interpret the input as the latent state at the initial time step, $x=x_0$. 
Conditioned on the initial latent state, we can calculate the predictive distribution $p(y|x, \phi)$ as
 \begin{equation}
 p(y|x, \phi) = \int p(y|x_T)    p(x_T|x, w_0) p(w_0|\phi) d w_0, x_T.  
 \label{eq:ctn_time_layer}
 \end{equation}
The transition kernel $p(x_T|x, w_0)$ is defined by the augmented dynamics, as discussed in Sec. \ref{subsec:pdssm_adf}, and the emission density $p(y|x_T)$ follows Eq. \ref{eq:emission}.
The mapping from
$x$ to $x_T$ can be interpreted as a deep stochastic layer. As the latent state is given, the filtering step of our algorithm becomes futile. 

 The dataset $\mathcal{D}=\{(x^n,y^n)\}_{n=1}^N$ consists of $N$ input-output tuples.  
 The likelihood term $p(\mathcal{D}|\phi)$ in Eq. \eqref{eq:type2_map} is given by
 \begin{equation}
     p(\mathcal{D}|\phi) = \prod_{n=1}^N p(y^n|x^n, \phi),
 \end{equation}
 where $p(y^n|x^n, \phi)$ follows Eq. \eqref{eq:ctn_time_layer}.
Similar models have also been developed in the context of continuous depth layers for neural ordinary differential equations (ODEs) \cite{chen18_neuralode} and stochastic differential equations (SDEs) \cite{look20_dinsde, tzen19_nsdelimit}.

\subsubsection{Datasets}
We use eight UCI datasets with varying input dimensionality and size that can be downloaded from \href{http://archive.ics.uci.edu/ml}{here}. 
These datasets  are used in prior art for benchmarking stochastic models such as probabilistic neural networks \cite{lobato15_bayesbackprob, gal16_dropout} or Gaussian processes \cite{lindinger2020beyond, salimbeni2017doubly}. 
We follow the experimental protocol as  defined in \cite{lobato15_bayesbackprob}. In short, we use 20 random splits. For each split, we use $90\%$ of the data for training and $10\%$ for testing. 
We follow \cite{look20_dinsde} for the design of the network architecture.  The mean/covariance functions are neural nets with one hidden layer and 40/10 hidden units. The observation function is a single linear layer. 
Similarly as \cite{look20_dinsde}, we add a residual connection to the transition density, i.e., we use $x_t+f(x_t, w_t)$ instead of $f(x_t, w_t)$ in Eq. \ref{eq:transition}.

\subsubsection{Baselines} 
We compare different variants  of our method ProDSSM and provide benchmarks against commonly used regression baselines.

\textit{i) ProDSSM variants:}
    \begin{itemize}
        \item Det. vs MC: We may approximate Eq. \ref{eq:ctn_time_layer} either via
         \textit{Monte Carlo} (MC) simulation or by using our \textit{Deterministic} (Det.) method that we introduced in Sec. \ref{subsec:pdssm_adf}. We vary the number of particles, i.e., MC simulations, during training and test time.
         \item Local vs. Global: In the local approach, the weights are resampled  at each time step. Contrarily, the weights are sampled  once at the initial time step and then kept constant throughout the remaining time steps in the global approach (see. Eq. \eqref{eq:local_vs_global}).
    \end{itemize}
    
    \textit{ii) DSSM \cite{look20_dinsde}:} 
    This method is equal to our contribution when removing the weight uncertainty. 
        
    \textit{iii) Dropout \cite{gal16_dropout}:} This method uses a single feed-forward neural net to predict the output, i.e., it does not rely on continuous depth layers.
    Stochasticity is introduced by applying a Bernoulli distributed masking scheme in all affine layers. 
    
    \textit{iv) DVI \cite{wu2018_deterministic}:} This method proposes a deterministic inference scheme for Bayesian neural nets. 
    Uncertainty is introduced by allowing for weight uncertainty over the neural net weights. 
    Similarly as Dropout, this method uses a feed-forward neural net.

\subsubsection{Results}
We report the \textit{Negative Log-Likelihood} (NLL) in Tab. \ref{tab:uci_res_nll}, and the \textit{Root Mean Squared Error} (RMSE) in Tab. \ref{tab:uci_res_rmse} in App. \ref{app:add_results}.

First, we compare the local and the global weight approach (see Sec. \ref{subsec:pdssm_uwp}) using our deterministic approximation scheme. 
For five datasets, the differences between both methods are less than one standard error. For the remaining three datasets, the global variant did not converge within the time limit of 72 hours\footnote{We use a time limit for the training runs in order to limit our carbon footprint.
This is motivated by the high computational cost of the deterministic approximation for the global weight setting. The time limit is a multiple of 24 and at least $10\times$ the runtime of the deterministic approximation for the local setting. For training, we use a NVIDIA Tesla V100 with 32GB.}, and is therefore outperformed by its local alternative. 

Next, we compare the local and global weight approach when using an MC approximation and varying the number of particles. 
We observe lower NLL and RMSE as we increase the number of particles. In order to achieve good predictive performance, a high number of particles is required. 
The local variant is in five datasets, while the global variant is only in three datasets among the best-performing methods.
We conjecture that the difference in performance can be attributed to
the higher gradient variance for the global variant, which makes training more difficult.

Using 128 MC samples and focusing on the local weight variant, MC sampling and our deterministic approximation perform en par except for the Naval dataset\footnote{There is little uncertainty in the  Naval dataset, and the better predictive performance of the MC variant can most likely be attributed to numerical issues.}.
However, it is important to note that our deterministic approximation is computationally more efficient, and restricting the MC approach to the same computational budget would result in approximately 12 samples, which is not sufficient for good performance.

Lastly, we compare our method against established baselines. ProDSSM with local weights is for five out of eight datasets among the best-performing methods in terms of NLL, thereby outperforming its competitors.

\begin{table*}[ht!]
	\caption{Negative log-likelihood for 8 datasets. We report average and standard error over 20 runs. Results marked by * did not converge within 72 hours. }
	\label{tab:uci_res_nll}
	\centering
	\resizebox{0.99\textwidth}{!}{
		\begin{tabular}{ l c  cc cc cc cc}
		\toprule
			& & Boston & Energy & Concrete & Wine Red & Kin8nm & Power & Naval & Protein \\
			\midrule
			Dropout&& 2.46(0.06) & 1.99(0.02) & {3.04(0.02)} & 0.93(0.01)& -0.95(0.01) &
			\textbf{2.80(0.01)}&-3.80(0.01)&2.89(0.00)\\
			DVI && {2.41(0.02)} & 1.01(0.06) & 3.06(0.01) & \textbf{0.90(0.01)} & {-1.13(0.00)} & \textbf{2.80(0.00)} & \textbf{-6.29(0.04)}& 2.85(0.00)\\
			DSSM && {2.37(0.03)} & 0.70(0.06) & \textbf{2.92(0.02)} & {0.93(0.02)} & {-1.22(0.00)} & \textbf{2.80(0.01)} & {-4.45(0.02)}& \textbf{2.76(0.01)}\\
			\midrule
			ProDSSM: MC, Local \\
			Train: \hspace{2.8mm}8 $\vert\vert$ Test:\hspace{1.4mm}32 && 2.42(0.03) & 0.47(0.03) & 3.02(0.02) & 0.96(0.01)&-1.25(0.00) & 2.85(0.01)& -5.88(0.09)& 2.86(0.01)\\
			Train: \hspace{2.8mm}8 $\vert\vert$ Test:128 && 2.41(0.02) & \textbf{0.44(0.03)} & 3.01(0.03) & 0.95(0.01)& -1.28(0.00) & 2.83(0.01)& -5.91(0.08) & 2.84(0.01)\\
			Train: \hspace{1.4mm}32 $\vert\vert$ Test:\hspace{1.4mm}32 && 2.38(0.03) & 0.47(0.06) & 3.06(0.03) & 0.95(0.01)&-1.27(0.01) & 2.82(0.01)& -6.08(0.07)& 2.81(0.01)\\
			Train:  \hspace{1.4mm}32 $\vert\vert$ Test:128 && 2.37(0.02) & \textbf{0.43(0.04)} & 2.99(0.01) & 0.93(0.01)& \textbf{-1.29(0.01)} & \textbf{2.79(0.01)}& -6.10(0.07)& \textbf{2.77(0.01)}\\
			Train: 128 $\vert\vert$ Test:\hspace{1.4mm}32 && 2.42(0.04) & \textbf{0.45(0.05)} & 3.09(0.04) & 0.96(0.01)&-1.26(0.01) & 2.83(0.01)& -6.15(0.07)& 2.83(0.01)\\
			Train: 128 $\vert\vert$ Test:128 && \textbf{2.36(0.03)} & \textbf{0.42(0.04)} & 3.00(0.03) & 0.93(0.01)& \textbf{-1.30(0.01)} & \textbf{2.79(0.01)}& -6.17(0.07)& \textbf{2.77(0.01)}\\
			\midrule
			ProDSSM: MC, Global\\
    		Train: \hspace{2.8mm}8 $\vert\vert$ Test:\hspace{1.4mm}32 && 2.49(0.02) & 0.56(0.03) & 3.08(0.02) & 0.96(0.01)&-1.22(0.01) & 2.85(0.01)& -6.16(0.05)& 2.89(0.01)\\
			Train: \hspace{2.8mm}8 $\vert\vert$ Test:128 && 2.46(0.02) & 0.54(0.03) & 3.06(0.01) & 0.94(0.01)&-1.24(0.01) & 2.83(0.01)& -6.19(0.05)& 2.87(0.01)\\
			Train: \hspace{1.4mm}32 $\vert\vert$ Test:\hspace{1.4mm}32 && 2.50(0.06) & 0.52(0.06) & 3.08(0.02) & 0.96(0.01)& -1.22(0.01) & 2.84(0.01)& -6.18(0.07)& 2.81(0.01)\\
			Train:  \hspace{1.4mm}32 $\vert\vert$ Test:128 && 2.44(0.05) & 0.50(0.06) & 3.03(0.02) & 0.93(0.01)& -1.25(0.01) & 2.81(0.01)& -6.22(0.07)& \textbf{2.77(0.01)}\\
			Train: 128 $\vert\vert$ Test:\hspace{1.4mm}32 && 2.44(0.04) & 0.54(0.05) & 3.10(0.04) & 0.97(0.02)& -1.22(0.01) & 2.83(0.01)& \textbf{-6.28(0.05)}& 2.82(0.01)\\
			Train: 128 $\vert\vert$ Test:128 && 2.41(0.04) & 0.50(0.05) & {3.03(0.02)} & 0.93(0.01)& -1.25(0.01) & \textbf{2.80(0.01)}& \textbf{-6.30(0.04)} & \textbf{2.77(0.01)}\\
			\midrule
			ProDSSM: Det., Local && \textbf{2.33(0.03)} & \textbf{0.43(0.04)} & 3.00(0.03) & 0.92(0.01)& \textbf{-1.30(0.00)} & \textbf{2.79(0.01)}& -5.52(0.03)& \textbf{2.76(0.01)}\\
			ProDSSM: Det., Global && \textbf{2.34(0.02)} & \textbf{0.44(0.03)} & 2.99(0.04) & 0.92(0.00) & -1.27(0.01)* & \textbf{2.79(0.01)}& -4.75(0.08)* & 2.82(0.01)*\\
			\bottomrule
		\end{tabular}
	}	
\end{table*}

\subsection{Filtering}
\label{subsec:experiments_filtering}
Next, we benchmark our new moment matching propagation scheme from Sec. \ref{subsec:pdssm_gf} for Gaussian filters on a standard filtering task. 

\subsubsection{Datasets}
In order to ensure that this experiment only evaluates the performance with respect to the filtering task, we use a two-step approach for creating data. In the first step, we create probabilistic ground-truth models; in the second step, we apply our newly created models in order to generate data for the filtering task.

\textit{Step 1:} We first train DSSM and our two ProDSSM variants on the kink dataset, which 
describes a non-linear dynamical system with varying emission noise $r=\{0.008, 0.08, 0.8\}$. 
See also Sec. \ref{subsec:experiments_dynamicalsystems} for more details.  

After this step, we obtained nine trained models with three different emission noise levels and with three different model variants.

\textit{Step 2:} For each trained model, we construct a new dataset by 
 simulating trajectories with a (Pro-)DSSM. 
Each trajectory has length $T=120$, and we simulate 10 trajectories per dataset. 
We evaluate the performance of different filtering methods on the NLL and RMSE of observing the true latent state on these nine newly created datasets.
The transition and emission functions in this experiment  are thereby fixed to the ground truth dynamics from Step 1. 

\subsubsection{Baselines}
We benchmark our filtering algorithm against two established baselines.

\textit{i) Unscented Kalman Filter} (UKF). 
This filter and our method share similarities as they are both instances of the Gaussian filter (see. Sec. \ref{subsec:background_filtering}). In contrast to our moment propagation approach, the intractable integrals are solved by using the unscented transform that is a numerical integration scheme \cite{saerkkae13_filtering}. 

\textit{ii) Neural Filter} (NF). 
In DSSM literature \cite{bayer2021_mind}, it is common practice to train a neural net based filter or smoother jointly with the generative model by maximizing ELBO.
During training, we fix the transition and emission function to the ground truth from Step 1.
We follow \cite{krishnan17_dks} for network design and 
use a recurrent neural net architecture that produces a sample $x_t$ at each time step $t$ as a function of the prior latent state $x_{t-1}$ and the observation $y_t$.

\subsubsection{Results}
We report results in Tab. \ref{tab:filtering}.
We observe for all methods that with increasing emission noise, it becomes more difficult to infer the latent distribution from the observations.
For deterministic weights, our method performs on par with UKF, while NF is outperformed for medium and higher noise levels.

 When switching to probabilistic weight modeling methods, the UKF has higher RMSE and NLL compared to our deterministic method for middle and high emission noise. 
Increasing the emission noise makes learning the dynamics more challenging and, as a result, leads to higher weight uncertainties.
We can also observe this behavior empirically:
  For low/middle/high observation noise, the average variance of the weights is $0.10/0.29/0.60$ for local weights and $0.05/0.26/0.49$ for global weights.
 As a consequence, the integration steps in the Gaussian filter become more difficult for increasing noise levels, and the performance of the UKF method deteriorates.
 In contrast, our newly introduced moment matching scheme performs well across the complete range of noise levels.
 
\begin{table}[t!]
\caption{NLL and MSE on a non-linear filtering dataset.
We report average and standard error over 10 runs.}
\label{tab:filtering}
\begin{center}
\resizebox{\columnwidth}{!}{\begin{tabular}{c l cccccc }
\toprule
 && \multicolumn{2}{c}{$r=0.008$}  
 & \multicolumn{2}{c}{$r=0.08$}  
 & \multicolumn{2}{c}{$r=0.8$}  
 \\
&& MSE & NLL & MSE & NLL & MSE & NLL\\
\midrule
\parbox[t]{0mm}{\multirow{3}{*}{\rotatebox[origin=c]{90}{Det.}}}  
&NF & \textbf{0.01(0.00)} & \textbf{-0.87(0.08)} & \textbf{0.08(0.01)} & {0.25(0.10)} & {0.73(0.19)} & {1.23(0.11)}\\
&UKF & \textbf{0.01(0.00)} & \textbf{-0.89(0.05)} & \textbf{0.07(0.00)} & \textbf{0.09(0.05)} & \textbf{0.49(0.08)} & \textbf{1.03(0.08)}
\\
&Ours & \textbf{0.01(0.00)} & \textbf{-0.88(0.04)} & \textbf{0.07(0.01)} & \textbf{0.10(0.04)}  & \textbf{0.46(0.07)} & \textbf{1.00(0.08)}\\
\midrule
\parbox[t]{0mm}{\multirow{2}{*}{\rotatebox[origin=c]{90}{Loc.}}}  
&UKF & \textbf{0.02(0.01)} & \textbf{-0.85(0.06)} & {0.12(0.02)} & {0.64(0.20)} & {1.35(0.15)} & {3.10(0.46)}
\\
&Ours & \textbf{0.01(0.00)} & \textbf{-0.90(0.02)} & \textbf{0.07(0.03)} & \textbf{0.08(0.03)}  & \textbf{0.44(0.05)} & \textbf{1.00(0.05)}\\
\midrule
\parbox[t]{0mm}{\multirow{2}{*}{\rotatebox[origin=c]{90}{Glob.}}}  
&UKF & \textbf{0.01(0.00)} & \textbf{-0.91(0.01)} & {0.27(0.06)} & {2.89(0.83)} & {1.18(0.25)} & {3.56(0.89)}
\\
&Ours & \textbf{0.01(0.00)} & \textbf{-0.89(0.02)} & \textbf{0.06(0.02)} & \textbf{0.12(0.03)}  & \textbf{0.48(0.04)} & \textbf{0.98(0.06)}\\
 \bottomrule
\end{tabular}}
\end{center}\end{table}

\subsection{Dynamical System Modeling}
\label{subsec:experiments_dynamicalsystems}

Our proposed model family, ProDSSM, is a natural choice for dynamical system modeling, where we aim to learn the underlying dynamics from a dataset $\mathcal{D}=\{Y^n\}_{n=1}^N$ consisting of $N$ trajectories.
For simplicity, we assume that each trajectory $Y^n=\{y_t^n\}_{t=1}^T$is of length $T$. 
Using the chain rule, the likelihood term $p(\mathcal{D}|\phi)$ in Eq. \eqref{eq:type2_map} can be written as
 \begin{equation}
     p(\mathcal{D}|\phi) = \prod_{n=1}^N 
     \prod_{t=1}^{T-1} p(y^n_{t+1}|y^n_{1:t}, \phi),
 \end{equation}
where we can approximate the predictive distribution $p(y^n_{t+1}|y^n_{1:t}, \phi)$ in a deterministic way as discussed in Sec. \ref{subsec:predictive_distribution}.

\subsubsection{Datasets}
We benchmark our method on two different datasets. 
The first dataset is a well-established learning task with synthetic non-linear dynamics, and the second dataset is a challenging real-world dataset.  

\textit{i) Kink} \cite{ialongo19_vcdt}:
We construct three datasets with varying degrees of difficulty by varying the emission noise level.
The transition density is given by $ \mathcal{N}(x_{t+1}|f_{kink}(x_t), 0.05^2)$ where $f_{kink}(x_t) = 0.8+(x_t+0.2)[1-5/(1+e^{-2x_t})]$ is the kink function.
The emission density is defined as $\mathcal{N}(y_t | x_t, r)$, where we vary $r$ between $\{0.008, 0.08, 0.8\}$. 
We simulate for each value of $r$ 10 trajectories of length $T=120$.  

We follow the experimental protocol as defined in \cite{lindinger22_laplacegp} and perform 10 training runs where each  run uses data from a single simulated trajectory only.
The mean function is realized with a neural net with one hidden layer and 50 hidden units, and the variance as a trainable constant.
For MC based ProDSSM variants, we use 64 samples during training. 
The cost of our deterministic approximation for the local approach is $\approx$50 samples. 

We compare the performance of the different methods with respect to epistemic uncertainty, i.e., parameter uncertainty, by evaluating if the learned transition model $p(x_{t+1}\vert x_t)$ covers the ground-truth dynamics.
In order to calculate NLL and MSE, we place 70 evaluation points on an equally spaced grid between the minimum and maximum latent state of the ground truth time series  and approximate for each point $x_t$ the mean $\mathbb{E}[x_t]=\int f(x_t, w_t) p(w_t) d w_t$ and variance $\text{Var}[x_t] = \int (f(x_t, w_t) - \mathbb{E}[x_t])^2 p(w_t) dw_t$ using 256 Monte Carlo samples. 

This dataset is commonly used for benchmarking GP based dynamical models \cite{ialongo19_vcdt, lindinger22_laplacegp}.
To the best of our knowledge, it has not been used in the context of DSSMs prior to this work.

\textit{ii) Mocap }: We follow \cite{yildiz2019_ode2vae} for preprocessing and designing the experimental setup.
The data is available \href{http://mocap.cs.cmu.edu/}{here}.
 It consists of 23 sequences from a single person. We use 16 sequences for training, 3 for validation, and 4 for testing. Each sequence consists of measurements from 50 different sensors. 
 We follow \cite{yildiz2019_ode2vae} for designing the network architecture and add a residual connection to the transition density, i.e., we use $x_t+f(x_t, w_t)$ instead of $f(x_t, w_t)$ in Eq. \ref{eq:transition}. 
 For MC based ProDSSM variants, we use 32 samples during training and 256 during testing. 
 The cost of our deterministic approximation for the local approach is approximately 24 samples. 
For numerical comparison, we compute NLL and MSE on the test sequences.

\subsubsection{Baselines}
We use the same ProDSSM variants as in our deep stochastic layer experiment (Sec. \ref{subsec:experiments_continuousdepthlayers}). 
Additionally, we compare against well-established baselines from GP and neural net based 
dynamical modeling literature. 

    \textit{i) VCDT  \cite{ialongo19_vcdt}:}
    This method relies on GPs to model a SSM. The distribution of the latent state is forward propagated via sampling. 
    Training is performed using doubly stochastic variational inference jointly over the GP posterior and the latent states.
    
    \textit{ii) Laplace GP \cite{lindinger22_laplacegp}:}
A GP based dynamical model that applies stochastic variational inference for the
Gaussian process posterior and the Laplace approximation over the latent states.

\textit{iii) ODE2VAE \cite{yildiz2019_ode2vae}:} The dynamics are modeled as latent neural ordinary differential equations. Stochasticity is introduced by accounting for uncertainty over the weights.
Contrary to our method, an additional neural net is used to approximate the latent distribution of the initial latent state, and the model does not account for transition noise.

\textit{iv) E-PAC-Bayes-Hybrid \cite{haussmann2021_pacsde}:}
The dynamics is modeled as a neural stochastic differential equation and accounts for aleatoric and epistemic uncertainty. 
Marginalization over the latent states and the weights is performed using Monte Carlo sampling.
This method focuses on integrating prior knowledge, either in the form of physics or by transfer learning across similar tasks, into the dynamics, 
hence the term \textit{Hybrid}. For the kink dataset, we reimplement this method without using prior knowledge.

\subsubsection{Results} 
\begin{figure}[h]
    \centering
    \includegraphics[width=.95\columnwidth ]{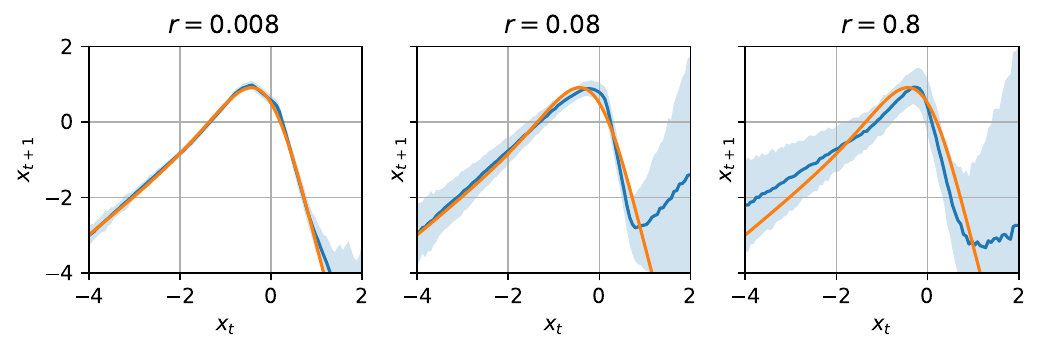}
    \caption{For increasing noise level $r$, we observe increased epistemic uncertainty. We visualize the true mean function $f(x_t)$ as an orange solid line. The blue solid line is the expected value of the learned mean function, and the shaded area represents 
    the 95\% confidence interval. 
    }\label{fig:kink}
\end{figure}

First, we analyze the results on the kink dataset.
We visualize the learned transition model of our model in Fig. \ref{fig:kink}. 
The confidence intervals capture the true transition function well, and the epistemic uncertainty increases with increasing noise levels.

We present the numerical results of this benchmark in Tab. \ref{tab:dynamical_modeling}.
For low  ($r=0.008$) and middle emission noise ($r=0.08$), all of our ProDSSM variants achieve on par performance with existing GP based dynamical models and outperform ODE2VAE.
For high emission noise ($r=0.08$), our ProDSSM variants perform significantly better than previous approaches.
The MC variants achieve for low and middle noise levels the same performance as the deterministic variants. 
As the noise is low, there is little function uncertainty, and few MC samples are sufficient for accurate approximations of the moments. 
If the emission noise is high, the marginalization over the latent states and the weights becomes more demanding, and the MC variant is outperformed by its deterministic counterpart.
Furthermore, we observe that for high observation noise, the local weight variant of our ProDSSM model achieves lower NLL than the global variant. 

We cannot report results for DSSM since this model does not account for epistemic uncertainty.

On the Mocap dataset, our best-performing ProDSSM variant from the previous experiments, which is the local weight variant together with the deterministic inference algorithm, is able to outperform all baselines. 
This is despite the fact that E-PAC-Bayes-Hybrid uses an additional dataset from another motion-capture task. 
Compared to the kink dataset, the differences between the MC and deterministic ProDSSM variants become more prominent: the Mocap dataset is high dimensional, and hence more MC samples are needed for accurate approximations. 
The ProDSSM variant with global weights and the deterministic inference was not able to converge within the time limit. 

\begin{table*}[t!]
	\caption{NLL and MSE for different dynamical system modeling tasks.
We report average and standard error over 10 runs.  Results marked by * did not converge within 120 hours. (NR=Not Reported, NA=Not Applicable)}
\label{tab:dynamical_modeling}
\begin{center}
\begin{tabular}{l cccccc | cc}
\toprule
 & \multicolumn{6}{c|}{Kink} & \multicolumn{2}{c}{Mocap}\\
 & \multicolumn{2}{c}{$r=0.008$}  
 & \multicolumn{2}{c}{$r=0.08$}  
 & \multicolumn{2}{c|}{$r=0.8$}  
 \\
& MSE & NLL & MSE & NLL & MSE & NLL & MSE & NLL\\
\midrule
VCDT &NR& \textbf{-1.53(0.31)} &NR& 1.10(0.72) &NR& 4.16(1.97) &NR&NR  \\
Laplace GP &NR& \textbf{-1.35(0.04)}&NR& {-0.36(0.08)}  &NR& 1.08(0.15) &NR&NR\\  
 ODE2VAE &0.01(0.00)&-0.07(0.46)& \textbf{0.02(0.00)}&0.29(0.43)&\textbf{0.22(0.05)}&4.11(1.42)&8.09(0.62)&NR\\
 E-PAC-Bayes-Hybrid  &0.02(0.00)&-0.67(0.07)&0.05(0.00)&0.47(0.11)&0.41(0.07)&1.13(0.12)&7.84(0.44)&253.64(20.01)   \\
DSSM &NA&NA&NA&NA&NA&NA& 7.87(0.69)&64.65(1.27) \\
\midrule
ProDSSM: MC, Local & \textbf{0.00(0.00)} & \textbf{-1.46(0.04)} &{0.04(0.01)}& \textbf{-0.44(0.06)} &0.28(0.05)& {0.82(0.11)} 
& 10.36(0.67)&74.74(1.68)  \\
ProDSSM: MC, Global &\textbf{0.00(0.00)}& \textbf{-1.50(0.07)} &{0.04(0.01)}& \textbf{-0.46(0.10)} &0.31(0.05)& {1.13(0.24)} 
& 10.65(1.25)&71.42(1.70) \\
ProDSSM: Det., Local &\textbf{0.00(0.00)}& \textbf{-1.50(0.03)} &{0.04(0.01)}& \textbf{-0.41(0.07)} &\textbf{0.22(0.04)}& \textbf{0.54(0.07)} & \textbf{6.98(0.17)}& \textbf{61.99(0.53)} \\
ProDSSM: Det., Global &\textbf{0.00(0.00)}& \textbf{-1.53(0.03)} &{0.03(0.01)}& \textbf{-0.47(0.07)} &\textbf{0.22(0.05)}& {0.72(0.17)}
& 21.29(0.82)* & 67.34(1.28)* \\
 \bottomrule
\end{tabular}
\end{center}\end{table*}

\subsection{Summary}
\label{subsec:experiments_summary}
Our experiments have demonstrated that our model family,  ProDSSM, performs favorably compared to state-of-the-art alternatives over a wide range of scenarios.
 Its benefits become especially pronounced when tackling complex datasets characterized by high noise levels or a high number of output dimensions.

First, we compare the local and global variants of our approach.
In the local variant, we resample the weight at each time step, while, for the global variant, we keep the weights fixed for the complete trajectory.
Independently of the chosen inference scheme, our experiments did not find a clear winner, provided that both variants converged.
However, the local variant is mathematically more convenient as it decorrelates subsequent time steps.
This property can be exploited for sample-free inference, where it results in a lower computational burden. 
Our empirical evidence confirms that this variant leads to more feasible solutions, whereas the global alternative is much slower and often did not converge in a reasonable amount of time.

Focusing on the local approach,
we can observe that our moment matching inference scheme outperforms its MC counterpart when using the same computational budget.
Disregarding runtime constraints, the MC variant still fails to surpass the performance of its deterministic alternative, indicating that (i) the Gaussian assumption is appropriate and (ii) the approximation error of our propagation scheme is negligible.

Despite the increased computational complexity of the global approach, we believe it warrants further exploration due to its ability to facilitate uncertainty decomposition \cite{depeweg2017_decomposition}, i.e., allowing for the separation of aleatoric and epistemic uncertainty. 
In contrast, the local approach does not support uncertainty decomposition, as both sources of uncertainty are intertwined at each time step. 
Additionally, the global approach could prove advantageous when transitioning from discrete to continuous dynamical systems, where achieving a parsimonious solution across different numerical solvers and step sizes is desirable.

\section{Conclusion}
\label{sec:conclusion}
In this work, we present ProDSSMs, a general framework for modeling unknown dynamical systems that respect epistemic and aleatoric uncertainty.
Inference for this model class is hard since we need to propagate the uncertainty over the neural network weights and of the latent states along a trajectory.
We address this challenge by introducing a novel inference scheme that exploits the internal structure of ProDSSMs and enjoys sample-free inference.
Our algorithm is general and can be applied to a variety of tasks and account for different weight sampling strategies.

In our experiments, we observe that our deterministic algorithm with local weights achieves better predictive performance in terms of lower NLL and MSE than its sampling-based counterpart under a fixed computational budget.
Compared to state-of-the-art alternatives, ProDSSM performs favorably over a wide range of scenarios.
The strengths of the method play out in particular on demanding datasets such as high-noise transition dynamics or high-dimensional outputs.

A  drawback of our algorithm is its reliance on the Gaussian assumption. A potential future research direction is the combination of our method with  Gaussian mixture filtering algorithms \cite{matti15_gmm_filter, ba06_gmm_filter}.  

\bibliographystyle{IEEEtran}
\bibliography{main}

\begin{thebibliography}{10}
\providecommand{\url}[1]{#1}
\csname url@samestyle\endcsname
\providecommand{\newblock}{\relax}
\providecommand{\bibinfo}[2]{#2}
\providecommand{\BIBentrySTDinterwordspacing}{\spaceskip=0pt\relax}
\providecommand{\BIBentryALTinterwordstretchfactor}{4}
\providecommand{\BIBentryALTinterwordspacing}{\spaceskip=\fontdimen2\font plus
\BIBentryALTinterwordstretchfactor\fontdimen3\font minus
  \fontdimen4\font\relax}
\providecommand{\BIBforeignlanguage}[2]{{%
\expandafter\ifx\csname l@#1\endcsname\relax
\typeout{** WARNING: IEEEtran.bst: No hyphenation pattern has been}%
\typeout{** loaded for the language `#1'. Using the pattern for}%
\typeout{** the default language instead.}%
\else
\language=\csname l@#1\endcsname
\fi
#2}}
\providecommand{\BIBdecl}{\relax}
\BIBdecl

\bibitem{kendall2017uncertainties}
A.~Kendall and Y.~Gal, ``{What Uncertainties Do We Need in Bayesian Deep
  Learning for Computer Vision?}'' \emph{Advances in neural information
  processing systems}, vol.~30, 2017.

\bibitem{depeweg2017_decomposition}
S.~Depeweg, J.~M. Hern{\'a}ndez-Lobato, F.~Doshi-Velez, and S.~Udluft,
  ``{Decomposition of Uncertainty in Bayesian Deep Learning for Efficient and
  Risk-sensitive Learning},'' in \emph{ICML}, 2017.

\bibitem{archer2015black}
E.~Archer, I.~M. Park, L.~Buesing, J.~Cunningham, and L.~Paninski, ``Black box
  variational inference for state space models,'' \emph{ICLR}, 2016.

\bibitem{karl2016deep}
M.~Karl, M.~Soelch, J.~Bayer, and P.~Van~der Smagt, ``{Deep Variational Bayes
  Filters: Unsupervised Learning of State Space Models from Raw Data},''
  \emph{ICLR}, 2017.

\bibitem{krishnan17_dks}
R.~G. Krishnan, U.~Shalit, and D.~Sontag, ``{Structured Inference Networks for
  Nonlinear State Space Models},'' in \emph{AAAI}, 2017.

\bibitem{yildiz2019_ode2vae}
C.~Yildiz, M.~Heinonen, and H.~Lahdesmaki, ``{ODE2VAE: Deep generative second
  order ODEs with Bayesian neural networks},'' in \emph{NeurIPS}, 2019.

\bibitem{dandekar2020bayesian}
R.~Dandekar, K.~Chung, V.~Dixit, M.~Tarek, A.~Garcia-Valadez, K.~V. Vemula, and
  C.~Rackauckas, ``{Bayesian Neural Ordinary Differential Equations},''
  \emph{arXiv preprint arXiv:2012.07244}, 2020.

\bibitem{iakovlev2022latent}
V.~Iakovlev, C.~Yildiz, M.~Heinonen, and H.~L{\"a}hdesm{\"a}ki, ``{Latent
  Neural ODEs with Sparse Bayesian Multiple Shooting},'' \emph{ICLR}, 2023.

\bibitem{depeweg2016learning}
S.~Depeweg, J.~M. Hern{\'a}ndez-Lobato, F.~Doshi-Velez, and S.~Udluft,
  ``{Learning and Policy Search in Stochastic Dynamical Systems with Bayesian
  Neural Networks},'' \emph{ICLR}, 2017.

\bibitem{ialongo19_vcdt}
A.~D. Ialongo, M.~Van Der~Wilk, J.~Hensman, and C.~E. Rasmussen, ``{Overcoming
  Mean-Field Approximations in Recurrent {G}aussian Process Models},'' in
  \emph{ICML}, 2019.

\bibitem{doerr18_prism}
A.~Doerr, C.~Daniel, M.~Schiegg, N.-T. Duy, S.~Schaal, M.~Toussaint, and
  T.~Sebastian, ``{Probabilistic Recurrent State-Space Models},'' in
  \emph{ICML}, 2018.

\bibitem{lindinger22_laplacegp}
J.~Lindinger, B.~Rakitsch, and C.~Lippert, ``{Laplace Approximated Gaussian
  Process State-Space Models},'' in \emph{UAI}, 2022.

\bibitem{haussmann2021_pacsde}
M.~Haussmann, S.~Gerwinn, A.~Look, B.~Rakitsch, and M.~Kandemir, ``{Learning
  Partially Known Stochastic Dynamics with Empirical PAC Bayes},'' in
  \emph{AISTATS}, 2021.

\bibitem{saerkkae13_filtering}
S.~S{\"a}rkk{\"a}, \emph{{Bayesian Filtering and Smoothing}}.\hskip 1em plus
  0.5em minus 0.4em\relax Cambridge University Press, 2013.

\bibitem{bayer2021_mind}
J.~Bayer, M.~Soelch, A.~Mirchev, B.~Kayalibay, and P.~van~der Smagt, ``{Mind
  the Gap when Conditioning Amortised Inference in Sequential Latent-Variable
  Models},'' in \emph{ICLR}, 2021.

\bibitem{bezenac20_nfkalman}
E.~de~B\'{e}zenac, S.~S. Rangapuram, K.~Benidis, M.~Bohlke-Schneider, R.~Kurle,
  L.~Stella, H.~Hasson, P.~Gallinari, and T.~Januschowski, ``{Normalizing
  Kalman Filters for Multivariate Time Series Analysis},'' in \emph{NeurIPS},
  2020.

\bibitem{brandt02_simulatedll}
M.~W. Brandt and P.~Santa-Clara, ``{Simulated Likelihood Estimation of
  Diffusions with an Application to Exchange Rate Dynamics in Incomplete
  Markets},'' \emph{Journal of Financial Economics}, vol.~63, no. 274, 2002.

\bibitem{petersen95_simulatedll}
A.~R. Pedersen, ``{A New Approach to Maximum Likelihood Estimation for
  Stochastic Differential Equations Based on Discrete Observations},''
  \emph{Scandinavian Journal of Statistics}, vol.~22, no.~1, 1995.

\bibitem{saerkkae15_inference}
S.~S\"{a}rkk\"{a}, J.~Hartikainen, I.~S. Mbalawata, and H.~Haario, ``{Posterior
  Inference on Parameters of Stochastic Differential Equations via Non-Linear
  Gaussian Filtering and Adaptive MCMC},'' \emph{Statistics and Computing},
  vol.~25, no.~2, 2015.

\bibitem{wu2018_deterministic}
A.~Wu, S.~Nowozin, E.~Meeds, R.~E. Turner, J.~M. Hernandez-Lobato, and A.~L.
  Gaunt, ``{Deterministic Variational Inference for Robust Bayesian Neural
  Networks},'' in \emph{ICLR}, 2019.

\bibitem{look2023_cheap}
A.~Look, B.~Rakitsch, M.~Kandemir, and J.~Peters, ``{Cheap and Deterministic
  Inference for Deep State-Space Models of Interacting Dynamical Systems},''
  \emph{TMLR}, 2023.

\bibitem{solin21_fpk}
A.~Solin, E.~Tamir, and P.~Verma, ``{Scalable Inference in SDEs by Direct
  Matching of the Fokker\textendash Planck\textendash Kolmogorov Equation},''
  in \emph{NeurIPS}, 2021.

\bibitem{look20_dinsde}
A.~Look, M.~Kandemir, B.~Rakitsch, and J.~Peters, ``{A Deterministic
  Approximation to Neural SDEs},'' \emph{IEEE TPAMI}, vol.~45, no.~4, 2023.

\bibitem{haussmann2021_bayesian}
M.~Haussmann, S.~Gerwinn, and M.~Kandemir, ``{Bayesian Evidential Deep Learning
  with {\{}PAC{\}} Regularization},'' in \emph{AABI}, 2021.

\bibitem{jazwinski1970_filtering}
A.~Jazwinski, \emph{{Stochastic Processes and Filtering Theory}}.\hskip 1em
  plus 0.5em minus 0.4em\relax Acad. Press, 1970.

\bibitem{chua18_pets}
{Chua, Kurtland and Calandra, Roberto and McAllister, Rowan and Levine,
  Sergey}, ``{Deep Reinforcement Learning in a Handful of Trials using
  Probabilistic Dynamics Models},'' in \emph{NeurIPS}, 2018.

\bibitem{liu1994_steinslemam}
J.~S. Liu, ``{Siegel's formula via Stein's identities},'' \emph{Statistics \&
  Probability Letters}, vol.~21, 1994.

\bibitem{murphy2013_machine}
K.~P. Murphy, \emph{{Machine Learning: A Probabilistic Perspective}}.\hskip 1em
  plus 0.5em minus 0.4em\relax MIT Press, 2013.

\bibitem{futami22_pbvi}
F.~Futami, T.~Iwata, N.~Ueda, I.~Sato, and M.~Sugiyama, ``{ Predictive
  variational Bayesian inference as risk-seeking optimization },'' in
  \emph{AISTATS}, 2022.

\bibitem{masegosa20_modelmiss}
A.~R. Masegosa, ``{Learning under Model Misspecification: Applications to
  Variational and Ensemble Methods},'' in \emph{NeurIPS}, 2020.

\bibitem{morningstar22a_pacm}
W.~R. Morningstar, A.~Alemi, and J.~V. Dillon, ``{PACm-Bayes: Narrowing the
  Empirical Risk Gap in the Misspecified Bayesian Regime},'' in \emph{AISTATS},
  2022.

\bibitem{maddox19_swag}
W.~J. Maddox, T.~Garipov, P.~Izmailov, D.~Vetrov, and A.~G. Wilson, ``{A Simple
  Baseline for Bayesian Uncertainty in Deep Learning},'' in \emph{NeurIPS},
  2019.

\bibitem{graves11_practicalvi}
A.~Graves, ``{Practical Variational Inference for Neural Networks},'' in
  \emph{NeurIPS}, 2011.

\bibitem{chung15_vrnn}
J.~Chung, K.~Kastner, L.~Dinh, K.~Goel, A.~C. Courville, and Y.~Bengio, ``{A
  Recurrent Latent Variable Model for Sequential Data},'' in \emph{NeurIPS},
  2015.

\bibitem{chen18_neuralode}
T.~Q. Chen, Y.~Rubanova, J.~Bettencourt, and D.~K. Duvenaud, ``{Neural Ordinary
  Differential Equations},'' in \emph{NeurIPS}, 2018.

\bibitem{tzen19_nsdelimit}
B.~Tzen and M.~Raginsky, ``{Neural Stochastic Differential Equations: Deep
  Latent Gaussian Models in the Diffusion Limit},'' \emph{ArXiv}, vol.
  abs/1905.09883, 2019.

\bibitem{lobato15_bayesbackprob}
J.~M. Hernandez-Lobato and R.~Adams, ``{Probabilistic Backpropagation for
  Scalable Learning of Bayesian Neural Networks},'' in \emph{ICML}, 2015.

\bibitem{gal16_dropout}
Y.~Gal and Z.~Ghahramani, ``{Dropout as a Bayesian Approximation: Representing
  Model Uncertainty in Deep Learning},'' in \emph{ICML}, 2016.

\bibitem{lindinger2020beyond}
J.~Lindinger, D.~Reeb, C.~Lippert, and B.~Rakitsch, ``{Beyond the Mean-Field:
  Structured Deep Gaussian Processes Improve the Predictive Uncertainties},''
  \emph{Advances in Neural Information Processing Systems}, vol.~33, pp.
  8498--8509, 2020.

\bibitem{salimbeni2017doubly}
H.~Salimbeni and M.~Deisenroth, ``{Doubly Stochastic Variational Inference for
  Deep Gaussian Processes},'' \emph{Advances in neural information processing
  systems}, vol.~30, 2017.

\bibitem{matti15_gmm_filter}
M.~Raitoharju, S.~Ali-Löytty, and R.~Piché, ``{Binomial Gaussian mixture
  filter},'' \emph{EURASIP Journal on Advances in Signal Processing}, vol.
  2015, 2015.

\bibitem{ba06_gmm_filter}
B.-N. Vo and W.-K. Ma, ``{The Gaussian Mixture Probability Hypothesis Density
  Filter},'' \emph{IEEE Transactions on Signal Processing}, vol.~54, no.~11,
  2006.

\bibitem{wick1950_isserlistheorem}
G.~C. Wick, ``{The Evaluation of the Collision Matrix},'' \emph{Phys. Rev.},
  vol.~80, 1950.

\end{thebibliography}
\begin{IEEEbiography}[{\includegraphics[width=1in,height=1.25in,clip,keepaspectratio]{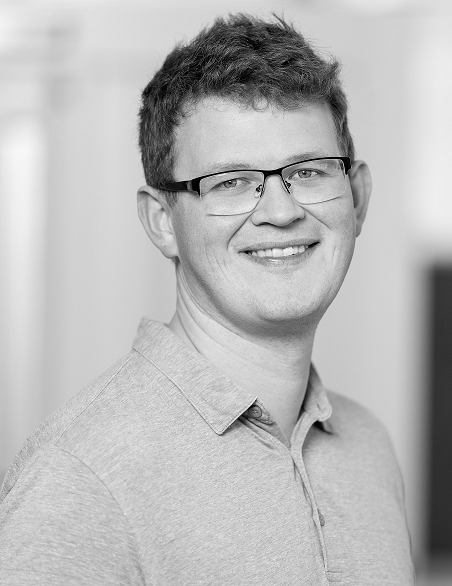}}]{Andreas Look}
 received his B.Sc. and M.Sc. from the University of Erlangen-Nuremberg in Power Engineering in 2014 and 2016, respectively. He had done research at the Institute of Fluid Mechanics and Hydraulic Machinery, University of Stuttgart, from 2016 until 2019. He joined the Bosch Center for Artificial Intelligence in 2019. His research interests are hybrid machine learning, differential equations, and optimization. \end{IEEEbiography}

\begin{IEEEbiography}[{\includegraphics[width=1in,height=1.25in,clip,keepaspectratio]{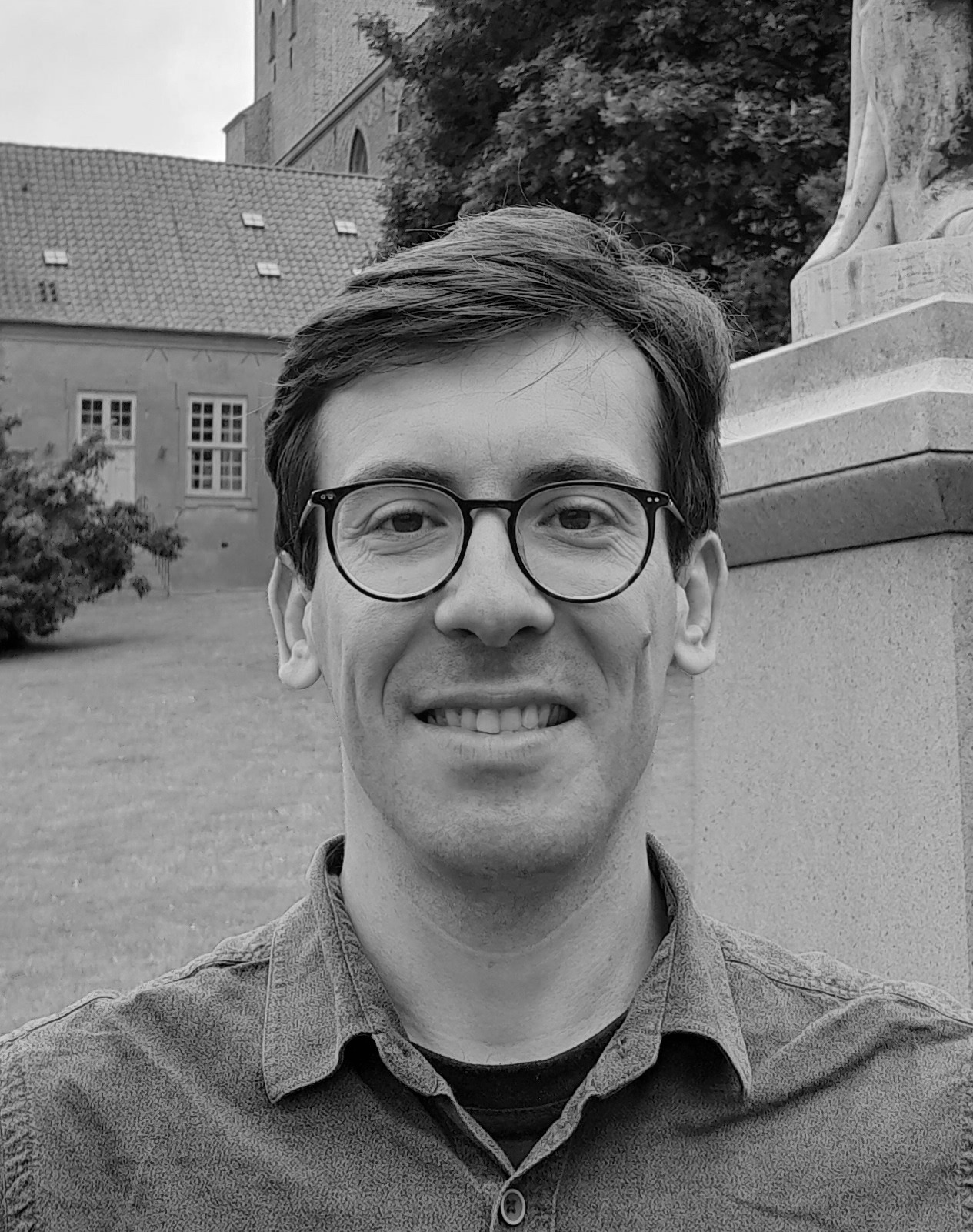}}]{Melih Kandemir} is an associate professor at the University of Southern Denmark (SDU), Department of Mathematics and Computer Science. Kandemir earned his PhD degree from Aalto University in 2013. He worked as a postdoctoral researcher at Heidelberg University, as an assistant professor at Ozyegin University in Istanbul, Turkey, and as a research group leader at Bosch Center for Artificial Intelligence. Kandemir pursues basic research on Bayesian inference and stochastic process modeling with deep neural nets with application to reinforcement learning and continual learning. Kandemir is an ELLIS member since 2021.
\end{IEEEbiography}

\begin{IEEEbiography}[{\includegraphics[width=1in,height=1.25in,clip,keepaspectratio]{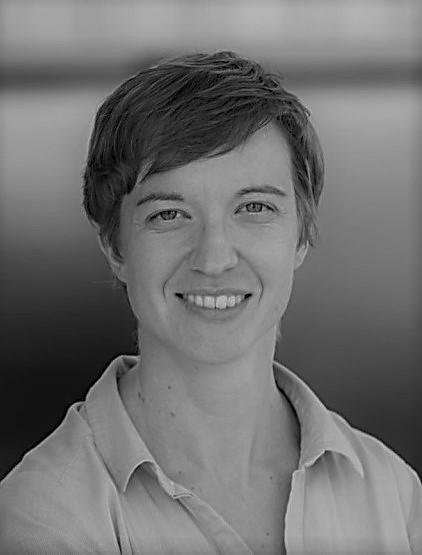}}]{Barbara Rakitsch} has been working as a research scientist at the Bosch Center for Artifical Intelligence in Renningen since 2017. Her interests lie in the area of Bayesian modeling with a focus on Gaussian processes and time-series data.
In 2014, she received her PhD in in probabilistic modeling for computational biology at the Max Planck Institute for Intelligent Systems in Tuebingen. Before joining Bosch, she worked on machine learning problems as a post-doc at the European Bioinformatics Institute in Cambridge and as a researcher in a cancer startup.
\end{IEEEbiography}

\begin{IEEEbiography}[{\includegraphics[width=1in,height=1.25in,clip,keepaspectratio]{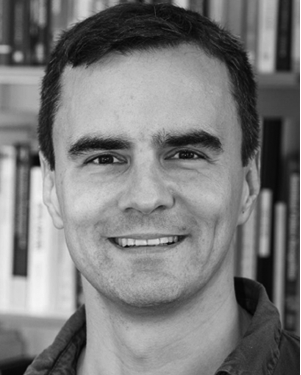}}]{Jan Peters}
is a full professor (W3) in intelligent autonomous systems, Computer Science Department, Technical University of Darmstadt, and at the same time a senior research scientist and group leader with the Max-Planck Institute for Intelligent Systems, where he heads the interdepartmental Robot Learning Group. He has received the Dick Volz Best 2007 US PhD Thesis Runner-Up Award, Robotics: Science \& Systems Early Career Spotlight, INNS Young Investigator Award,and IEEE Robotics \& Automation Society’s Early Career Award as well as numerous best paper awards. In 2015, he received an ERC Starting Grant and in 2019, he was appointed as an IEEE fellow.
\end{IEEEbiography}

\clearpage
\appendices

\section{Moments of the Linear Layer}
\label{app:moments_linear_layer}

A linear layer applies an affine transformation 
\begin{align}
     U({z}^l_t) =  
     \begin{bmatrix}
    A^l_t {x}_t^l + b^l_t\\
     w_t^l
     \end{bmatrix},  
 \end{align}
 where the transformation matrix $A^l_t\in \mathbb{R}^{D_x^{l+1} \times D_x^{l}}$ and bias $b^l_t\in \mathbb{R}^{D_x^{l+1}}$ are both part of weights $(A^l_t,b_t^l) \in w_t^l$. As the cross-covariance matrix $\Sigma_t^{l,xw}$ is non-zero  for global weights, the  transformation matrix $A^l_t$, bias $b_t^l$, and state $x_t^l$ are assumed to be jointly normally distributed. Contrarily, for the local case, the cross-covariance matrix $\Sigma_t^{l,xw}$ is zero, i.e., weights and states are uncorrelated.
The $i$-th output ${x}_{t,i}^{l+1} \in \mathbb{R}$ of the affine transformation is calculated as
\begin{equation}
    {x}^{l+1}_{t, i} =
    \sum_{m=1}^{D_x^l} a^l_{t,im} {x}^l_{t,m} + b^l_{t,i},
\end{equation}
where $a_{t,ij}^l$ is $(i,j)$-th entry in $A_t^l$ and $b^l_{t,i}$ the $i$-th entry in $b^l_t$. 
Given the above update rule, we can calculate the output moments of the affine transformation as
\begin{align}
    m^{l+1,x}_{t,i} &= \sum_{m=1}^{D_x^l}
    \mathbb{E}[a_{t,im}^l {x}^l_{t,m}] + \mathbb{E}[b_{t,i}^l], \label{eq:affine_mean_x}\\
     \Sigma^{l+1, x}_{t, ij}
     &=
    \sum_{m,n=1}^{D_x^l}
    \text{Cov}[a^l_{t,im} {x}^l_{t,m}, a^l_{t,jn} {x}^l_{t,n}] 
    +\nonumber\\
    &~~~~~~\sum_{m=1}^{D_x^l}
    \text{Cov}[b^l_{t,i}, a^l_{t,jm} {x}^l_{t,m}] 
    +\nonumber\\
    &~~~~~~\sum_{m=1}^{D_x^l}
    \text{Cov}[a^l_{t,im} {x}^l_{t,m}, b^l_{t,j}] 
    +
    \text{Cov}[b^l_{t,i}, b^l_{t,j}], \label{eq:affine_cov_x}\\
    \Sigma^{l+1, xw}_{t. ij}&=
    \sum_{m=1}^{D_x^l}
    \text{Cov}[a^l_{t,im} {x}^l_{t,m}, w_{t,j}^l] 
    +
    \text{Cov}[b_{t,i}^l, w_{t,j}^l], \label{eq:affine_cov_xw}
\end{align}
which is a direct result of the linearity of the $\text{Cov}[\bullet, \bullet]$ operator.
In order to compute the above  moments, we need to  calculate the 
moments of a product of correlated normal variables, $\mathbb{E}[a_{t,im}^l{x}_{t,m}^l], \text{Cov}[
    a_{t,im}^l{x}_{t,m}^l,
    a_{t,jn}^l{x}_{t,n}^l],$ and $\text{Cov}[
    a_{t,im}^l{x}_{t,m}^l,
    w_{t,j}^l]$.
The output moments of a product of correlated normal variables are calculated as
\begin{align}
    \mathbb{E}[a_{t,im}^l{x}_{t,m}^l]
    &\!=\!
    \text{Cov}[a_{t,im}^l, {x}_{t,m}^l]
    +\nonumber\\
    &~~~~~\mathbb{E}[a_{t,im}^l]
    \mathbb{E}[{x}_{t,m}^l],
    \label{eq:prod_mean}\\
    \text{Cov}[
    a_{t,im}^l{x}_{t,m}^l,
    a_{t,jn}^l{x}_{t,n}^l]
    &\!=\!
    \text{Cov}[a_{t,im}^l, a_{t,jn}^l]
    \text{Cov}[{x}_{t,m}^l,
    {x}_{t,n}^l]
    +\nonumber\\
    &~~~~\text{Cov}[a_{t,im}^l, a_{t,jn}^l]
    \mathbb{E}[{x}_{t,m}^l]
    \mathbb{E}[{x}_{t,n}^l]
    +\nonumber\\
    &~~~~\text{Cov}[{x}_{t,m}^l, {x}_{t,n}^l]
    \mathbb{E}[a_{t,im}^l]
    \mathbb{E}[a_{t,jn}^l]
    +\nonumber\\
    &~~~~\text{Cov}[
     a_{t,im}^l, {x}_{t,n}^l]
    \text{Cov}[
     {x}_{t,m}^l, a_{t,jn}^l]
     +\nonumber\\
     &~~~~\text{Cov}[
     a_{t,im}^l, {x}_{t,n}^l]
     \mathbb{E}[ {x}_{t,m}^l]
     \mathbb{E}[ a_{t,jn}^l]
     +\nonumber\\
    &~~~~\text{Cov}[
     {x}_{t,m}^l, a_{t,jn}^l]
     \mathbb{E}[a_{t,im}^l]
     \mathbb{E}[ {x}_{t,n}^l],
     \label{eq:prod_var}\\
     \text{Cov}[
    a_{t,im}^l{x}_{t,m}^l,
    w_{t,j}^l]
    &\!=\!
     \text{Cov}[
    a_{t,im}^l,
    w_{t,j}^l]\mathbb{E}[{x}_{t,m}^l]+
    \nonumber\\
     &~~~~~\text{Cov}[
    {x}_{t,m}^l,
    w_{t,j}^l]\mathbb{E}[a_{t,im}^l].
    \label{eq:prod_covar}
\end{align}
The above results are exact and hold for both local and global weights, as long as $x_t^l$ and $w_t^l$ follow a normal distribution.   
In the case of local weights 
the cross-covariance terms $\text{Cov}[a_{t,im}^l, a_{t,jn}^l]$ and $\text{Cov}[{x}_{t,m}^l, a_{t,jn}^l]$ are zero, as the weights are iid. 
Setting these cross-covariance terms in Eq. \ref{eq:prod_var} to zero recovers the result from \cite{wu2018_deterministic}.  

The expectation $\mathbb{E}[a_{t,im}^l{x}_{t,m}^l]$ follows straightforwardly from the definition of the covariance, i.e. 
$\text{Cov}[a_{t,im}^l, {x}_{t,m}^l] = \mathbb{E}[a_{t,im}^l{x}_{t,m}^l] - \mathbb{E}[a_{t,im}^l]
    \mathbb{E}[{x}_{t,m}^l]$.
Contrarily, computing $\text{Cov}[
    a_{t,im}^l{x}_{t,m}^l,
    a_{t,jn}^l{x}_{t,n}^l]$ is more sophisticated. 
In order to avoid cluttered notation we omit in the derivation the time and layer index. 
The arguments $a_{im}, x_m, a_{jn},  x_n$ are jointly Gaussian distributed
\begin{align}
    \begin{pmatrix}
    a_{im}\\
    x_{m}\\
    a_{jn}\\
    x_n
    \end{pmatrix}
    \sim
    \mathcal{N}
    \left(
    m
    ,
    \Sigma
    \right),
\end{align}
with mean and covariance
\begin{align}   
    m&=\begin{pmatrix}
    m_{a_{im}}\\
    m_{x_{m}}\\
    m_{a_{jn}}\\
    m_{x_{n}}
    \end{pmatrix},\\
    \Sigma&=
    \begin{pmatrix}
    \Sigma_{a_{im},a_{im}} & \Sigma_{a_{im}, x_{m}} & \Sigma_{a_{im}, a_{jn}} & \Sigma_{a_{im}, x_n}\\
    \Sigma_{x_{m},a_{im}} & \Sigma_{x_{m}, x_{m}} & \Sigma_{x_{m}, a_{jn}} & \Sigma_{x_{m}, x_n}\\
    \Sigma_{a_{jn},a_{im}} & \Sigma_{a_{jn}, x_{m}} & \Sigma_{a_{jn}, a_{jn}} & \Sigma_{a_{jn}, x_n}\\
    \Sigma_{x_n,a_{im}} & \Sigma_{x_n, x_{m}} & \Sigma_{x_n, a_{jn}} & \Sigma_{x_n, x_n}\\
    \end{pmatrix}.
\end{align}
We first calculate the expectation of the product of four Gaussian random variables
\begin{align}
        \mathbb{E}[a_{im}x_{m}a_{jn}x_n]\!=\!
 \mathbb{E}[
 &(a_{im}-m_{a_{im}}\!+\!m_{a_{im}})  
 (x_{m}-m_{x_{m}}\!+\!m_{x_{m}})\nonumber\\
 &(a_{jn}-m_{a_{jn}}\!+\!m_{a_{jn}})
 (x_{n}-m_{x_{n}}\!+\!m_{x_{n}})]\nonumber\\
 =\mathbb{E}[
 &(\bar{a}_{im}\!+\!m_{a_{im}})  
 (\bar{x}_{m}\!+\!m_{x_{m}})\nonumber\\
 &(\bar{a}_{jn}\!+\!m_{a_{jn}})
 (\bar{x}_{n}\!+\!m_{x_{n}})],
\end{align}
where the accent $\bar{x}$ denotes the centered version of the random variable $x$. We  execute the product and arrive at 
\begin{align}
        \mathbb{E}[a_{im}x_{m}a_{jn}x_n]\!=\!
    &m_{a_{im}} \!m_{x_{m}} \!m_{a_{jn}} \!m_{x_{m}} \!+\!
    \cancel{m_{a_{im}}\!m_{x_{m}} \! m_{a_{jn}} \mathbb{E}[\bar{x_{n}}]} \!+\!\nonumber\\
    &\cancel{m_{a_{im}}\!m_{x_{m}}\! m_{x_{m}}\!  \mathbb{E}[\bar{a_{jn}}]} \!+\!
    \cancel{m_{a_{im}}\!m_{a_{jn}}\! m_{x_{m}} \! \mathbb{E}[\bar{x_{m}}]} +\nonumber\\
    &\cancel{m_{x_{m}}\! m_{a_{jn}}\! m_{x_{m}} \! \mathbb{E}[\bar{a_{im}}]} \!+\!
    m_{a_{im}}\! m_{x_{m}}  \!\mathbb{E}[\bar{a_{jn}}  \bar{x_{n}}] \!+\!\nonumber\\ 
    &m_{a_{im}} m_{a_{jn}} \mathbb{E}[\bar{x_{m}}   \bar{x_{n}}] \!+\!
    m_{a_{im}} m_{x_n}    \mathbb{E}[\bar{x_{m}}   \bar{a_{jn}}] \!+\! \nonumber\\
    &m_{x_{m}}  m_{a_{jn}} \mathbb{E}[\bar{a_{im}}  \bar{x_{n}}] \!+\! 
    m_{x_{m}}  m_{x_n}    \mathbb{E}[\bar{a_{im}}  \bar{a_{jn}}] \!+\! \nonumber\\
    &m_{a_{jn}} m_{x_n}    \mathbb{E}[\bar{a_{im}}  \bar{x_{m}}] \!+ \!
    \cancel{m_{a_{im}} \mathbb{E}[\bar{x_{m}}  \bar{a_{jn}} \bar{x_{n}}]} +\nonumber\\
    &\cancel{m_{x_{m}}  \mathbb{E}[\bar{a_{im}} \bar{a_{jn}} \bar{x_{n}}]} \!+\!
    \cancel{m_{a_{jn}} \mathbb{E}[\bar{a_{im}} \bar{x_{m}}  \bar{x_{n}}]} +\nonumber\\
    &\cancel{m_{x_n}    \mathbb{E}[\bar{a_{im}} \bar{x_{m}}  \bar{a_{jn}}]} \!+\!
    \mathbb{E}[\bar{a_{im}}\bar{x_{m}}\bar{a_{jn}}\bar{x_{n}}] .
\end{align}
Due to Isserlis` theorem \cite{wick1950_isserlistheorem}, any odd central moment of a product of centered Gaussian random variables is zero. 
As a consequence, the expectations involving the products
of one/three centered Gaussians are zero. In order to 
calculate $\mathbb{E}[\bar{a_{im}}\bar{x_{m}}\bar{a_{jn}}\bar{x_{n}}]$ we make
once more use of Isserlis`theorem and arrive at
\begin{align}
    \mathbb{E}[\bar{a_{im}}\bar{x_{m}}\bar{a_{jn}}\bar{x_{n}}]
    =
    &\mathbb{E}[\bar{a_{im}} \bar{x_{m}}]\mathbb{E}[\bar{a_{jn}} \bar{x_{n}}]+\nonumber\\
    &\mathbb{E}[\bar{a_{im}} \bar{a_{jn}}]\mathbb{E}[\bar{x_{m}} \bar{x_{n}}]+\nonumber\\
    &\mathbb{E}[\bar{a_{im}} \bar{x_{n}}]\mathbb{E}[\bar{x_{m}} \bar{a_{jn}}].
\end{align}

\begin{table*}[t!]
	\caption{RMSE for 8 datasets. We report average and standard error over 20 runs. Results marked by * did not converge within 72 hours. (NR=Not Reported) }
	\label{tab:uci_res_rmse}
	\centering
	\resizebox{0.99\textwidth}{!}{
		\begin{tabular}{ l c  cc cc cc cc}
		\toprule
			& & Boston & Energy & Concrete & Wine Red & Kin8nm & Power & Naval & Protein \\
			\midrule
    		Dropout  && \textbf{2.97(0.19)} & 1.66(0.04) & 5.23(0.12) &\textbf{0.62(0.01)}& {0.10(0.00)} & \textbf{4.02(0.04)}& \textbf{0.01(0.00)}&\textbf{4.36(0.01)}\\
			DVI && NR & NR & NR & NR & NR & NR & NR & NR\\
			DSSM && {3.26(0.15)} & 0.87(0.13) & \textbf{5.12(0.09)} & \textbf{0.63(0.00)} & {0.08(0.00)} & {4.07(0.03)} & \textbf{0.01(0.00)}& 4.45(0.00)\\
			\midrule
			ProDSSM: MC, Local\\
			Train: \hspace{2.8mm}8 $\vert\vert$ Test:\hspace{1.4mm}32 && 3.17(0.14) & 0.43(0.01) & 5.48(0.10) & 0.64(0.00)& \textbf{0.07(0.00)} & 4.14(0.03)& \textbf{0.01(0.00)}& 4.63(0.02)\\
			Train: \hspace{2.8mm}8 $\vert\vert$ Test:128 && \textbf{3.14(0.13)} & \textbf{0.42(0.03)} & 5.36(0.11) & 0.64(0.00)& \textbf{0.07(0.00)} & 4.07(0.03)&\textbf{0.01(0.00)}& 4.56(0.02)\\
			Train: \hspace{1.4mm}32 $\vert\vert$ Test:\hspace{1.4mm}32 && \textbf{3.11(0.13)} & \textbf{0.41(0.01)} & 5.48(0.11) & \textbf{0.63(0.01)}& \textbf{0.07(0.00)} & 4.04(0.03)&\textbf{0.01(0.00)}& 4.46(0.01)\\
			Train:  \hspace{1.4mm}32 $\vert\vert$ Test:128 && \textbf{3.11(0.13)} & \textbf{0.41(0.01)} & 5.43(0.11) & \textbf{0.63(0.00)} & \textbf{0.07(0.00)} & \textbf{4.00(0.03)} &\textbf{0.01(0.00)}& 4.39(0.01)\\
			Train: 128 $\vert\vert$ Test:\hspace{1.4mm}32 && \textbf{3.05(0.12)} & \textbf{0.41(0.01)} & \textbf{5.21(0.08)} & 0.64(0.01)& \textbf{0.07(0.00)} & 4.04(0.03)&\textbf{0.01(0.00)}& 4.44(0.02)\\
			Train: 128 $\vert\vert$ Test:128 && \textbf{3.04(0.12)} & \textbf{0.41(0.01)} & \textbf{5.18(0.09)} & \textbf{0.63(0.00)} & \textbf{0.07(0.00)} & \textbf{4.00(0.03)} &\textbf{0.01(0.00)}& \textbf{4.37(0.02)}\\
			\midrule
			ProDSSM: MC, Global\\
    		Train: \hspace{2.8mm}8 $\vert\vert$ Test:\hspace{1.4mm}32 && 3.39(0.13) & 0.47(0.01) & 5.66(1.00) & 0.64(0.00)& \textbf{0.07(0.00)} & 4.15(0.03)&\textbf{0.01(0.00)}& 4.70(0.04)\\
			Train: \hspace{2.8mm}8 $\vert\vert$ Test:128 && 3.27(0.12) & 0.46(0.01) & 5.60(0.10) & \textbf{0.63(0.00)}& \textbf{0.07(0.00)} & 4.09(0.03)&\textbf{0.01(0.00)}& 4.62(0.03)\\
			Train: \hspace{1.4mm}32 $\vert\vert$ Test:\hspace{1.4mm}32 && 3.17(0.14) & 0.45(0.02) & 5.50(0.10) & 0.64(0.00)& \textbf{0.07(0.00)} & 4.09(0.03)&\textbf{0.01(0.00)}& 4.44(0.01)\\
			Train:  \hspace{1.4mm}32 $\vert\vert$ Test:128 && \textbf{3.15(0.13)} & 0.44(0.02) & 5.44(0.10) & \textbf{0.63(0.00)}& \textbf{0.07(0.00)} & 4.04(0.03)&\textbf{0.01(0.00)}& 4.39(0.01)\\
			Train: 128 $\vert\vert$ Test:\hspace{1.4mm}32 && \textbf{3.16(0.12)} & 0.46(0.01) & 5.53(0.07) & 0.64(0.00)& \textbf{0.07(0.00)} & 4.05(0.03)&\textbf{0.01(0.00)}& 4.40(0.01)\\
			Train: 128 $\vert\vert$ Test:128 && \textbf{3.14(0.12)} & 0.45(0.01) & 5.46(0.08) & \textbf{0.63(0.00)} & \textbf{0.07(0.00)} & \textbf{4.01(0.03)}&\textbf{0.01(0.00)}& \textbf{4.36(0.02)}\\
			\midrule
			ProDSSM: Det., Local && \textbf{2.99(0.13)} & \textbf{0.41(0.01)} & 5.24(0.12) & \textbf{0.63(0.00)} & \textbf{0.07(0.00)} & \textbf{3.99(0.03)} & \textbf{0.01(0.00)}& \textbf{4.35(0.02)}\\
			ProDSSM: Det., Global && \textbf{3.05(0.11)} & \textbf{0.42(0.02)} & 5.24(0.14) & \textbf{0.63(0.00)}& \textbf{0.07(0.00)}* & \textbf{4.01(0.03)}&  \textbf{0.01(0.00)*} & 4.57(0.03)*\\
			\bottomrule
		\end{tabular}
	}	
\end{table*}

Plugging everything together, we arrive at a tractable expression for the expectation
of four Gaussian random variables
\begin{align}
        \mathbb{E}[a_{im}x_{m}a_{jn}x_n]\!=\!
      &m_{a_{im}} m_{x_{m}} m_{a_{jn}}  m_{x_{n}}+
    \Sigma_{{a_{im}}, {x_{m}}}\Sigma_{{a_{jn}}, {x_{n}}}+\nonumber\\
    &\Sigma_{{a_{im}}, {x_{m}}}m_{a_{jn}} m_{x_{n}}+
    \Sigma_{{a_{jn}}, {x_{n}}}m_{a_{im}} m_{x_{m}}+\nonumber\\
    &\Sigma_{{a_{im}}, {a_{jn}}}\Sigma_{{x_{m}}, {x_{n}}}+
    \Sigma_{{a_{im}}, {a_{jn}}}m_{x_{m}} m_{x_{n}}+\nonumber\\
    &\Sigma_{{x_{m}},  {x_{n}}}m_{a_{im}} m_{a_{jn}} +
    \Sigma_{{a_{im}}, {x_{n}}}\Sigma_{x_m, {a_{jn}}}+\nonumber\\
    &\Sigma_{{a_{im}}, {x_{n}}}m_{x_{m}} m_{a_{jn}}+
    \Sigma_{{x_{m}}, {a_{jn}}}m_{a_{im}} m_{x_{n}}.
\end{align}
Given the above result, we can calculate
the covariance as
\begin{align}
    \text{Cov}[{a_{im}}{x_{m}},{a_{jn}}{x_{n}}]
    &=
    \mathbb{E}[{a_{im}}{x_{m}}{a_{jn}}{x_{n}}] \!-\!\mathbb{E}[{a_{im}}{x_{m}}]\mathbb{E}[{a_{jn}}{x_{n}}]\nonumber\\
    &=
    \Sigma_{{a_{im}}, {a_{jn}}} \Sigma_{{x_{m}}, {x_{n}}}+
    \Sigma_{{a_{im}}, {a_{jn}}} m_{x_{m}} m_{x_{n}}+\nonumber\\
    &~~~~~\Sigma_{{x_{m}},  {x_{n}}}  m_{a_{im}} m_{a_{jn}}+
    \Sigma_{{a_{im}}, {x_{n}}}  \Sigma_{{x_{m}}, {a_{jn}}}+\nonumber\\
    &~~~~~\Sigma_{{a_{im}}, {x_{n}}}  m_{x_{m}} m_{a_{jn}}+
    \Sigma_{{x_{m}},  {a_{jn}}} m_{a_{im}} m_{x_{n}}.
    \label{eq:app_cov_final_result}
\end{align}
We obtain the result for $\text{Cov}[a_{t,im}^l x_{t,m}^l, w_{t,j}^l]$ by setting 
${x_{n}}=1$ and ${a_{jn}}=w_j$ in Eq. \ref{eq:app_cov_final_result}
\begin{align}
    \text{Cov}[{a_{im}}{x_{m}},{w_{j}}]
    &=
    \cancel{\Sigma_{{a_{im}}, {w_{j}}} \Sigma_{{x_{m}}, 1}}+
    \Sigma_{{a_{im}}, {w_{j}}} m_{x_{m}} +\nonumber\\
    &~~~~~\cancel{\Sigma_{{x_{m}},  1}  m_{a_{im}} m_{a_{j}}}+
    \cancel{\Sigma_{{a_{im}}, 1}  \Sigma_{{x_{m}}, {w_{j}}}}+\nonumber\\
    &~~~~~\cancel{\Sigma_{{a_{im}}, 1}  m_{x_{m}} m_{w_{j}}}+
    \Sigma_{{x_{m}},  {w_{j}}} m_{a_{im}}\nonumber\\
     &= \Sigma_{{a_{im}}, {w_{j}}} m_{x_{m}}+
    \Sigma_{{x_{m}},  {w_{j}}} m_{a_{im}},
    \label{eq:app_crosscov_final_result}
\end{align}
as the cross-covariance between a random variable is zero.

\section{Similarities between ELBO and Predictive Variational
Bayesian Inference}
\label{app:background_bnn}


The classical Bayesian formalism defines a prior $p(w|\phi)$ with hyperparameters $\phi$ over the weights $w\in \mathbb{R}^{D_w}$
and a likelihood $p(\mathcal{D}|w)$ of observing the data $\mathcal{D}$.  The posterior $p(w|\mathcal{D})$ is the quantity of interest. During posterior inference, the hyperparameters $\phi$ of the prior are kept constant. As an analytical solution to the posterior is intractable, either \textit{Markov Chain Monte Carlo} (MCMC) \cite{maddox19_swag} or \textit{Variational Inference} (VI) \cite{graves11_practicalvi} is used. VI introduces an approximate posterior $q(w)$ and maximizes the \textit{Evidence Lower Bound
} (ELBO)
\begin{equation}
    \mathbb{E}_{q(w)}[\log p(\mathcal{D}|w)] -
    \text{KL}(q(w)|p(w|\phi)).
    \label{eq:elbo}
\end{equation}
Commonly, the approximate posterior is modeled as a Gaussian distribution \cite{wu2018_deterministic}. 
The KL-divergence between two Gaussians $q=\mathcal{N}(m^w, \Sigma^w)$ and $p=\mathcal{N}(m_p^w, \Sigma_p^w)$ with dimensionality $D_w$ is available in closed form as
\begin{align}
    \text{KL}(q|p)&=  \frac{1}{2} 
    \big[ 
    \log \frac{\Sigma_p^w}{\Sigma^w} - D + \text{Tr}((\Sigma_p^w)^{-1}\Sigma^w)\nonumber
    \\
    &~~~~+ (m_p^w-m^w)^T (\Sigma_p^w)^{-1}(m_p^w-m^w)
    \big],
\end{align}
where $m_i$ denotes the $i$-th entry of the mean vector of $q$.
We further assume that $q$ is modeled with a diagonal covariance and the $i$-th entry of the diagonal is denoted with
 $\Sigma_{ii}$. 
For the case of a standard normal prior $p(w|\phi)=\mathcal{N}(0,{I})$, the KL-divergence between the prior and approximate posterior takes the below form
\begin{align}
    \text{KL}(q|p)=
    \frac{1}{2} \sum_{i=1}^{D_w} -\log \Sigma_{ii}^w + \Sigma_{ii}^w + (m_i^w)^2 -const.,
\end{align}
which is equivalent to the negative hyper-prior in Eq. \ref{eq:hyper-prior}. We summarize that the ELBO is equivalent to our proposed training objective in Eq. \eqref{eq:type2_map} up to the position of the logarithm in the likelihood term.


\section{Additional results}
\label{app:add_results}
We present additional results for the UCI regression task. In Tab. \ref{tab:uci_res_rmse}, we show the RMSE values for our method and various baselines.

\end{document}